\documentclass[10pt,twocolumn,letterpaper]{article}

\usepackage[pagenumbers]{cvpr} 

\usepackage[dvipsnames]{xcolor}
\newcommand{\red}[1]{{\color{Red}#1}}
\newcommand{\green}[1]{{\color{Green} #1}}

\newcommand{\common}[1]{{\color[HTML]{0173b2} #1}}
\newcommand{\private}[1]{{\color[HTML]{de8f05} #1}}

\usepackage{epsfig}
\usepackage{amsmath}
\usepackage{amssymb}
\usepackage{booktabs}

\usepackage{amsmath,amsfonts,bm}

\def\eqref#1{equation~\ref{#1}}

\def\1{\bm{1}}

\DeclareMathAlphabet{\mathsfit}{\encodingdefault}{\sfdefault}{m}{sl}
\SetMathAlphabet{\mathsfit}{bold}{\encodingdefault}{\sfdefault}{bx}{n}

\def\gG{{\mathcal{G}}}

\def\gX{{\mathcal{X}}}
\def\gY{{\mathcal{Y}}}
\def\gZ{{\mathcal{Z}}}

\newcommand{\E}{\mathbb{E}}

\DeclareMathOperator*{\argmax}{arg\,max}

\usepackage{subcaption, multirow, array, tabularx}
\usepackage{algpseudocode}
\usepackage{adjustbox}
\usepackage{arydshln}
\usepackage{pifont}
\usepackage{capt-of}
\usepackage{comment}
\usepackage{mathtools, empheq}
\usepackage{enumitem}
\usepackage{makecell}
\usepackage{dsfont}
\usepackage[dvipsnames]{xcolor}

\newtheorem{lemma*}{Lemma}

\newtheorem{theorem*}{Theorem}

\newcommand{\cmark}{\ding{51}} 
\newcommand{\xmark}{\ding{55}} 
\newcommand{\rpm}{\raisebox{.2ex}{$\scriptstyle\pm$}}
\newcolumntype{P}[1]{>{\centering\arraybackslash}p{#1}}

\algnewcommand\algorithmicforeach{\textbf{for each}}
\algdef{S}[FOR]{ForEach}[1]{\algorithmicforeach\ #1\ \algorithmicdo}

\definecolor{cvprblue}{rgb}{0.21,0.49,0.74}
\usepackage[pagebackref,breaklinks,colorlinks,citecolor=cvprblue]{hyperref}
\usepackage[accsupp]{axessibility}

\def\paperID{14187} 
\def\confName{CVPR}
\def\confYear{2024}

\title{Universal Semi-Supervised Domain Adaptation \\ by Mitigating Common-Class Bias}

\author{Wenyu Zhang$^1$, Qingmu Liu$^2$\thanks{Contributed to this work while interning at A*STAR.}, Felix Ong Wei Cong$^{2*}$, Mohamed Ragab$^{1,3}$, Chuan-Sheng Foo$^{1,3}$\\
$^1$Institute for Infocomm Research (I$^\text{2}$R), Agency for Science, Technology and Research (A*STAR)\\
$^2$National University of Singapore (NUS)\\
$^3$Centre for Frontier AI Research (CFAR), Agency for Science, Technology and Research (A*STAR)\\
}

\begin{document}
\maketitle
\begin{abstract}
Domain adaptation is a critical task in machine learning that aims to improve model performance on a target domain by leveraging knowledge from a related source domain.  
In this work, we introduce Universal Semi-Supervised Domain Adaptation (UniSSDA), a practical yet challenging setting where the target domain is partially labeled, and the source and target label space may not strictly match. UniSSDA is at the intersection of Universal Domain Adaptation (UniDA) and Semi-Supervised Domain Adaptation (SSDA): the UniDA setting does not allow for fine-grained categorization of target private classes not represented in the source domain, while SSDA focuses on the restricted closed-set setting where source and target label spaces match exactly.
Existing UniDA and SSDA methods are susceptible to common-class bias in UniSSDA settings, where models overfit to data distributions of classes common to both domains at the expense of private classes. 
We propose a new prior-guided pseudo-label refinement strategy to reduce the reinforcement of common-class bias due to pseudo-labeling, a common label propagation strategy in domain adaptation.
We demonstrate the effectiveness of the proposed strategy on benchmark datasets Office-Home, DomainNet, and VisDA. The proposed strategy attains the best performance across UniSSDA adaptation settings and establishes a new baseline for UniSSDA.
\end{abstract}

\section{Introduction}
\label{sec: introduction}

Domain adaptation (DA) is a critical task in machine learning, where models are adapted from a source domain to perform well on a different target domain. Conventionally, DA works focus on the closed-set adaptation setting, where the source and target label space match exactly, to address covariate shift and potential label distribution shift \cite{Ganin2016DANN, zhao2018adversarial, Sun2016DeepCORAL, Wang2018DeepVD}. 
More recently, the generalized problem in unsupervised DA termed `Universal DA' (UniDA) \cite{you2019uan, chen2022maths, zhu2023uniam, saito2021ovanet} aims to bridge the gap between labeled source data and unlabeled target data, and considers challenging settings where there may be label space shift between the two domains. This means that, under UniDA, there may be source private and target private classes besides common classes shared across the two domains, as in the open-set, partial-set, and open-partial settings. The universal setup handles practical scenarios where different object classes are present in different environments, or when new task objectives require collecting new target domain object classes not present in the source domain. However, since the UniDA setting assumes target data is unlabeled, all target private classes are categorized under a single `unknown' class. Unsupervised domain adaptation is also known to be prone to negative transfer as target classes can be mapped to incorrect source classes \cite{zhao2019invariant, Li2020RethinkingDM}.

\begin{figure}[tb]
    \centering
    \includegraphics[width=0.9\columnwidth]{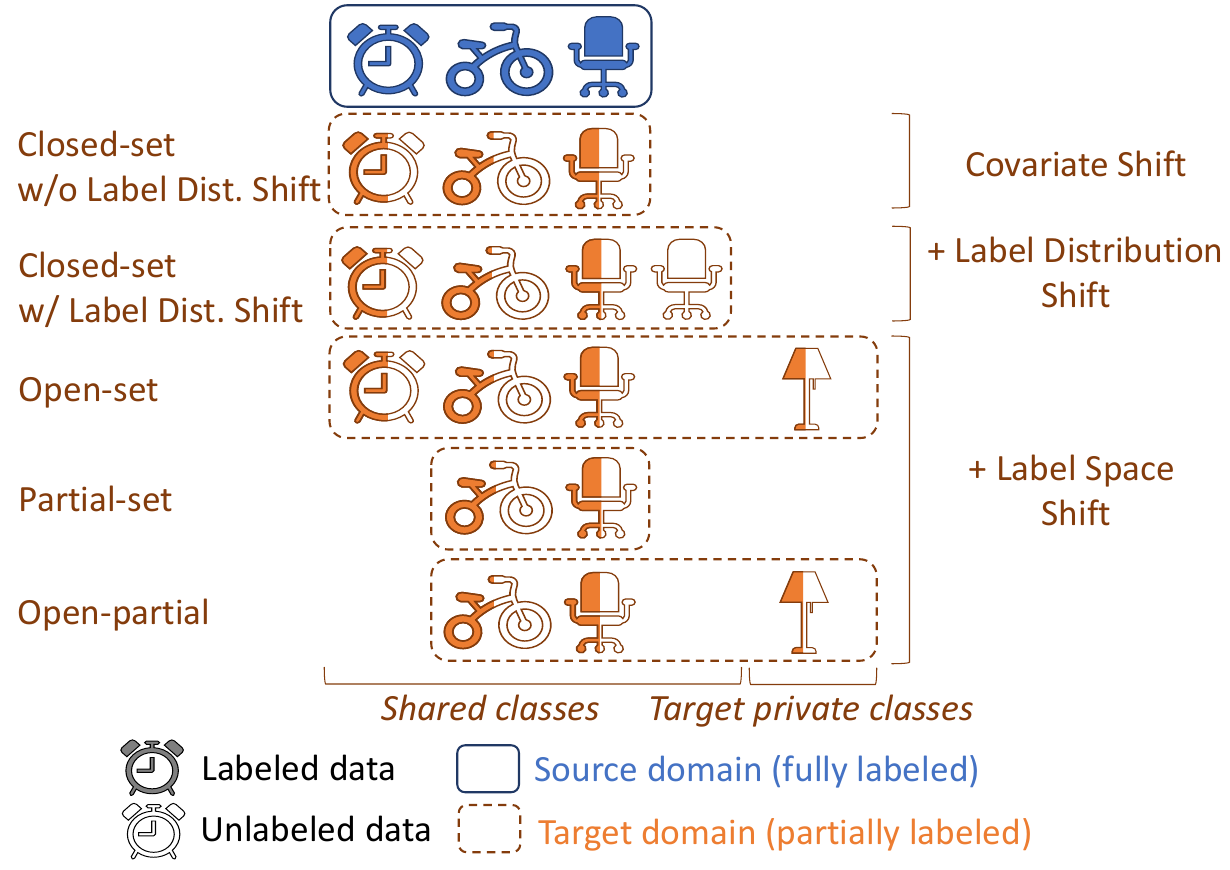}
    \vspace{-2mm}
    \caption{Adaptation settings in Universal SSDA. Existing SSDA works conventionally focus on closed-set settings.}
    \label{fig: adaptation_setting}
    \vspace{-4mm}
\end{figure}

In this work, we introduce a new setting to include a small number of labeled target samples in UniDA such that methods can provide fine-grained classification of target private class samples and better leverage target information that may be available. While the related semi-supervised domain adaptation (SSDA) setting also allows for partially labeled target data and offers a practical balance between data annotation and performance, works have been restricted to the closed-set setting \cite{Singh2021CLDA, mishra2021surprisingly, li2021cdac}. We thus propose the new generalized setting at the intersection of UniDA and SSDA as `Universal SSDA' (UniSSDA). 
In UniSSDA, methods need to improve adaptation performance regardless of the source and target label space as illustrated in Figure~\ref{fig: adaptation_setting}.
To the best of our knowledge, our work is the first study on UniSSDA.

\begin{figure}[tb]
     \centering
     \begin{subfigure}[b]{0.49\columnwidth}
         \centering
         \includegraphics[width=\textwidth]{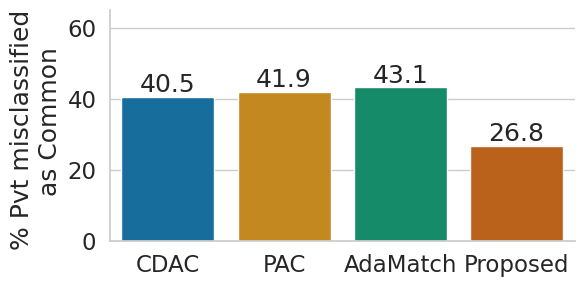}
         \caption{Comparison with SSDA}
         \label{fig: misclf_openpartial_trg_private_dm_ssda}
     \end{subfigure}
     \begin{subfigure}[b]{0.49\columnwidth}
         \centering
         \includegraphics[width=\textwidth]{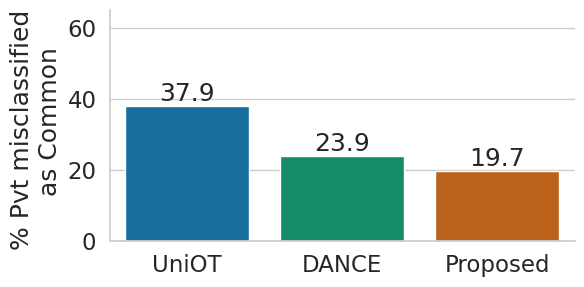}
         \caption{Comparison with UniDA}
         \label{fig: misclf_openpartial_trg_private_dm_unida}
     \end{subfigure}
     
    \vspace{-2mm}
    \caption{(a) and (b) show the percentage of samples in target private classes ($\mathcal{Y}_T \cap \mathcal{Y}_S^c$) misclassified as common classes ($\mathcal{Y}_T \cap \mathcal{Y}_S$) under open-partial setting, demonstrating the effect of common-class bias on existing SSDA and UniDA methods. (a) is implemented on DomainNet-126 $C\rightarrow P$ with ResNet-34 backbone. (b) is implemented on DomainNet-345 $C\rightarrow P$ with frozen ViT foundation model encoder and learnable classifier.}
    \label{fig: misclf_diagnostics_dm}
    \vspace{-4mm}
\end{figure}

\begin{figure}[tb]
     \centering
         \centering
         \includegraphics[width=\columnwidth]{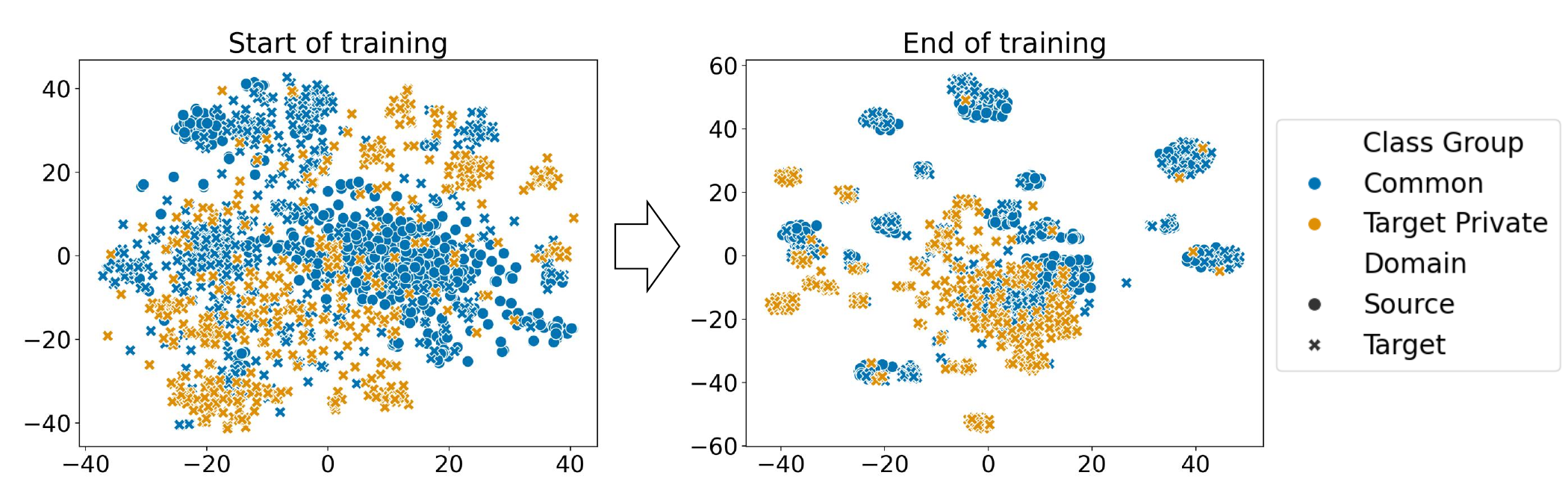}
     
    \vspace{-2mm}
    \caption{T-SNE visualization of \common{common} versus \private{target private} class samples shows negative transfer due to common-class bias. Incorrect mapping between common and target private classes can persist and be reinforced by naive pseudo-labeling. Example taken from DomainNet-126 $C\rightarrow P$ in open-partial setting, with visualization on 15 shared and 15 target private classes.}
    \label{fig: tsne}
    \vspace{-4mm}
\end{figure}

A key challenge in UniSSDA is that due to the abundance of labeled source data, model learning is naturally biased towards fitting the source distribution. This overfitting is especially detrimental to the learning of target private classes in UniSSDA settings, since these classes are not represented at all in the source domain. We refer to this bias as the `common-class bias', where model learning focuses on classes common to both domains at the expense of private classes. From Figure~\ref{fig: misclf_diagnostics_dm}, we observe that existing SSDA and UniDA methods are vulnerable to the common-class bias in a UniSSDA open-partial setting on DomainNet, and misclassify as much as 40\% target private samples as belonging to common classes.

We hypothesize that the propagation of biased label information to the unlabeled target samples resulted in the eventual negative transfer observed in target private classes. To gain further insight into this issue, we study the pseudo-labeling component commonly used in DA methods \cite{li2021cdac, yoon2022s3d, yan2022mcl, li2021ECACL, yang2020decota}, as naive pseudo-labeling tends to reinforce pre-existing biases as the model iteratively fits to incorrect pseudo-labels during training. 

Through t-SNE visualizations of the feature space, Figure~\ref{fig: tsne} illustrates the effects of the common-class bias when training utilizes a naive pseudo-labeling strategy where the class with the highest confidence score is assigned as the pseudo-label, and pseudo-labeled samples are selected by a confidence threshold. Target private classes are incorrectly mapped to common classes, and this incorrect mapping is reinforced during training. In this work, we propose a pseudo-label refinement strategy to mitigate common-class bias. We introduce prior-guidance to directly reweigh and refine the predictions on unlabeled target samples. The strategy is simple to implement and effective, and the refined pseudo-labels can be readily incorporated into existing adaptation algorithms. 

Our key contributions in this work are:
\begin{itemize}
    \item We are the first to study Universal SSDA (UniSSDA), a new setup at the intersection of UniDA and SSDA. Unlike UniDA, UniSSDA allows fine-grained classification of target private classes and is able to leverage available target label information. UniSSDA can also be viewed as a practical generalization of SSDA that removes the restriction for source and target label space to strictly match. 
    \item Experimental evaluations find that existing SSDA and UniDA methods do not consistently perform well in UniSSDA settings and are susceptible to common-class bias. This highlights the need to develop new approaches to address UniSSDA.
    \item We propose a prior-guided pseudo-label refinement strategy to reduce the reinforcement of common-class bias due to target pseudo-labeling, a common label propagation strategy in DA problems.
    \item We demonstrate the performance of the proposed strategy on 3 domain adaptation image classification datasets. The proposed strategy establishes a new UniSSDA baseline for future research work in this area.
\end{itemize}
\section{Related Works}
\label{sec: related works}

\subsection{Semi-Supervised Domain Adaptation}
\label{sec: ssda methods}

Existing works in SSDA aim to learn domain invariant representations and to mine intrinsic target domain structures. Methods focused on domain invariance align features \cite{cheng2014manifolds, li2021lirr, Kim2020AttractPA, yao2015sdasl, Singh2021CLDA, yoon2022s3d, yan2022mcl, li2021ECACL}, label distributions \cite{berthelot2021adamatch} or hypotheses \cite{ngo2021collaborative, kim2022dark, daume2010frustratingly, kumar2010coregularization, li2021lirr} between domains to transfer knowledge from source to target domain. \cite{cheng2014manifolds} directly maps source to target data manifold by learning a linear transformation. Following works in unsupervised domain adaptation \cite{Wang2018DeepVD, Ganin2016DANN, Sun2016DeepCORAL}, many SSDA methods reduce feature distribution mismatch by adversarial learning or minimizing a domain discrepancy measure \cite{li2021lirr, Kim2020AttractPA, yao2015sdasl, Singh2021CLDA, yoon2022s3d, yan2022mcl, li2021ECACL}. CLDA~\cite{Singh2021CLDA} uses contrastive learning and ECACL~\cite{li2021ECACL} uses a triplet loss to simultaneously pull same-class samples together and push different-class samples apart.

Since both labeled and unlabeled target samples are available during training, methods employ strategies including those from semi-supervised learning \cite{yang2022surveyssl} and self-supervised learning \cite{ohri2021surveyselfsl} to mine structures and information from target data. PAC~\cite{mishra2021surprisingly} uses rotation prediction as a pre-training task, and \cite{li2020metalearning} adds meta-learning to an existing SSDA method to find better model initializations. Several methods apply clustering objectives on the target domain to learn class-discriminative target features \cite{li2021cdac, donahue2013instance, saito2019minimax, yao2015sdasl, yoon2022s3d, yan2022mcl}. Sample similarity for clustering can be determined based on pseudo-label \cite{yoon2022s3d}, inter-sample distance \cite{yao2015sdasl}, or a given similarity graph \cite{donahue2013instance}. Some methods encourage prediction consistency of samples under different augmentations so as to find a smooth data manifold and to learn compact target clusters \cite{berthelot2021adamatch, mishra2021surprisingly, li2021cdac, Singh2021CLDA, yan2022mcl, li2021ECACL}. 

To propagate label information to unlabeled samples, methods typically use pseudo-labeling. BiAT~\cite{jiang2020biat} and S$^3$D \cite{yoon2022s3d} generate intermediate styles to bridge the domain gap between source and target domains to facilitate label propagation. 
Some methods use uncertainty measures, such as confidence and entropy, for data selection to improve pseudo-labeling accuracy \cite{li2021cdac, yoon2022s3d, yan2022mcl, li2021ECACL, yang2020decota}. However, pseudo-labeling risks reinforcing pre-existing source-induced bias, and we propose pseudo-label refinement strategies to mitigate such bias using duo classifiers. DST~\cite{chen2022DST}, a semi-supervised learning method, also utilizes an additional classifier, but DST cannot be applied to frozen foundation model feature extractors, and DST's main classifier may not fully represent the data distribution as it is not trained on any unlabeled samples.

\subsection{Universal Domain Adaptation}
\label{sec: universal da}

UniDA comprises settings where the source and/or target domain can have private classes, and assumes all target samples to be unlabeled. Similar to SSDA methods in Section~\ref{sec: ssda methods}, UniDA methods exploit intrinsic structures in target data by discovering clusters \cite{li2021dcc, saito2020dance, chang2022uniot}. UniDA methods also aim to learn domain invariant representations through domain alignment. MATHS \cite{chen2022maths} uses contrastive learning between mutual nearest neighbor samples for domain alignment, and UniAM \cite{zhu2023uniam} achieves domain-wise and category-wise alignment by feature and attention matching between domains in vision transformers. A key challenge in UniDA is to distinguish between common and private class samples such that alignment is only performed on the former, so as to avoid negative transfer. 
OVANet~\cite{saito2021ovanet} trains a one-vs-all classifier for each class to estimate the inter-class distance as a threshold to distinguish between the two class groups. Other methods use uncertainty measures such as entropy, confidence and consistency \cite{saito2020dance, chang2022uniot, singh2021selection, fu2020cmu, you2019uan}, sample similarity between domains \cite{you2019uan, singh2021selection}, or outlier detection on the logits \cite{chen2022maths} to detect target private class samples. A common assumption in UniDA is that the target private class samples are predicted with higher uncertainty because there are no labeled data from target private classes for supervised training. The assumption is not applicable in UniSSDA. A recent study finds that the supervised baseline is competitive with or outperforms existing UniDA methods on foundation models \cite{deng2023uniood}, hence we include foundation models in our experiments for a more comprehensive evaluation.
\section{UniSSDA Preliminaries}
\label{sec: preliminaries}

We denote the input and output space as $\gX$ and $\gY$. Distribution shift occurs when the source and target distribution differ (i.e. $p_S(x, y) \neq p_T(x, y)$), for instance in covariate shift (i.e. $p_S(x) \neq p_T(x)$) and label shift (i.e. $p_S(y) \neq p_T(y)$). Same as other domain adaptation tasks \cite{Ganin2016DANN,li2021cdac,you2019uan}, UniSSDA assumes the presence of covariate shift. In existing SSDA works, methods and evaluations are focused on the closed-set setting where source and target domain have the same label space (i.e. $\mathcal{Y_S} = \mathcal{Y_T}$), and label shift is restricted to label distribution shift where $\mathcal{Y_S} = \mathcal{Y_T}$ and $\exists y \in \mathcal{Y_S} \hspace{1mm} p_S(y) \neq p_T(y)$.

Label space shift involves a change in label set (i.e. $\mathcal{Y_S} \neq \mathcal{Y_T}$) \cite{you2019uan}, and UniSSDA methods need to be effective in these more challenging types of domain shift. To the best of our knowledge, we are the first to study UniSSDA. We consider the adaptation settings in Figure~\ref{fig: adaptation_setting}:
\begin{itemize}
    \item Closed-set with no label distribution shift (i.e. $\mathcal{Y_S} = \mathcal{Y_T}$, $\forall y \in \mathcal{Y_S} \hspace{1mm} p_S(y) = p_T(y)$);
    \item Closed-set with label distribution shift (i.e. $\mathcal{Y_S} = \mathcal{Y_T}$, $\exists y \in \mathcal{Y_S} \hspace{1mm} p_S(y) \neq p_T(y)$);
    \item Open-set: Source label space is a proper subset of target label space (i.e. $\mathcal{Y}_S \subset \mathcal{Y}_T$);
    \item Partial-set: Target label space is a proper subset of source label space (i.e. $\mathcal{Y}_T \subset \mathcal{Y}_S$);
    \item Open-partial: Source and target label space intersect but neither is a proper subset of the other (i.e. $\mathcal{Y}_S \cap \mathcal{Y}_T \neq \emptyset$ and $\mathcal{Y}_S \cap \mathcal{Y}_T^c \neq \emptyset$ and $\mathcal{Y}_T \cap \mathcal{Y}_S^c \neq \emptyset$).
\end{itemize}
Classes can be categorized into 3 groups based on domain label space membership: common ($\mathcal{Y}_T \cap \mathcal{Y}_S$), source private ($\mathcal{Y}_T^c \cap \mathcal{Y}_S$), and target private ($\mathcal{Y}_T \cap \mathcal{Y}_S^c$).
\section{Methodology}
\label{sec: proposed method}

\begin{figure}[tb]
     \centering
     \begin{subfigure}[b]{0.49\columnwidth}
         \centering
         \includegraphics[width=\textwidth]{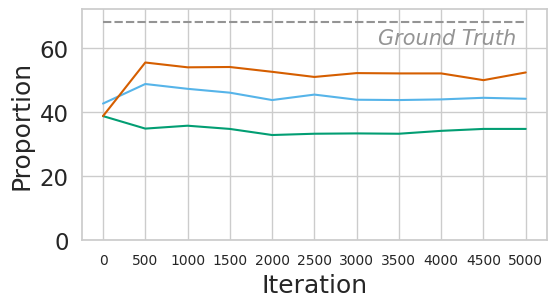}
         \caption{Proportion of predictions in \\ target private class}
         \label{fig: pct_trg_pvt}
     \end{subfigure}
     \begin{subfigure}[b]{0.49\columnwidth}
         \centering
         \includegraphics[width=\textwidth]{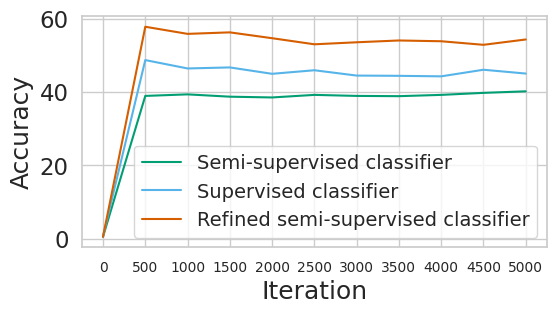}
         \caption{Target private class prediction accuracy}
         \label{fig: acc_trg_pvt}
     \end{subfigure}

    \vspace{-2mm}
    \caption{The semi-supervised classifier trained with naive pseudo-labels is more vulnerable to common-class bias than the supervised classifier is. Using supervised classifier outputs as priors to refine the pseudo-labels significantly improves the performance of the resulting semi-supervised classifier. Example taken from DomainNet $C\rightarrow P$ in open-partial setting.}
    \label{fig: classifiers}
    \vspace{-4mm}
\end{figure}

\begin{figure}[tb]
    \centering
    \includegraphics[width=0.5\textwidth]{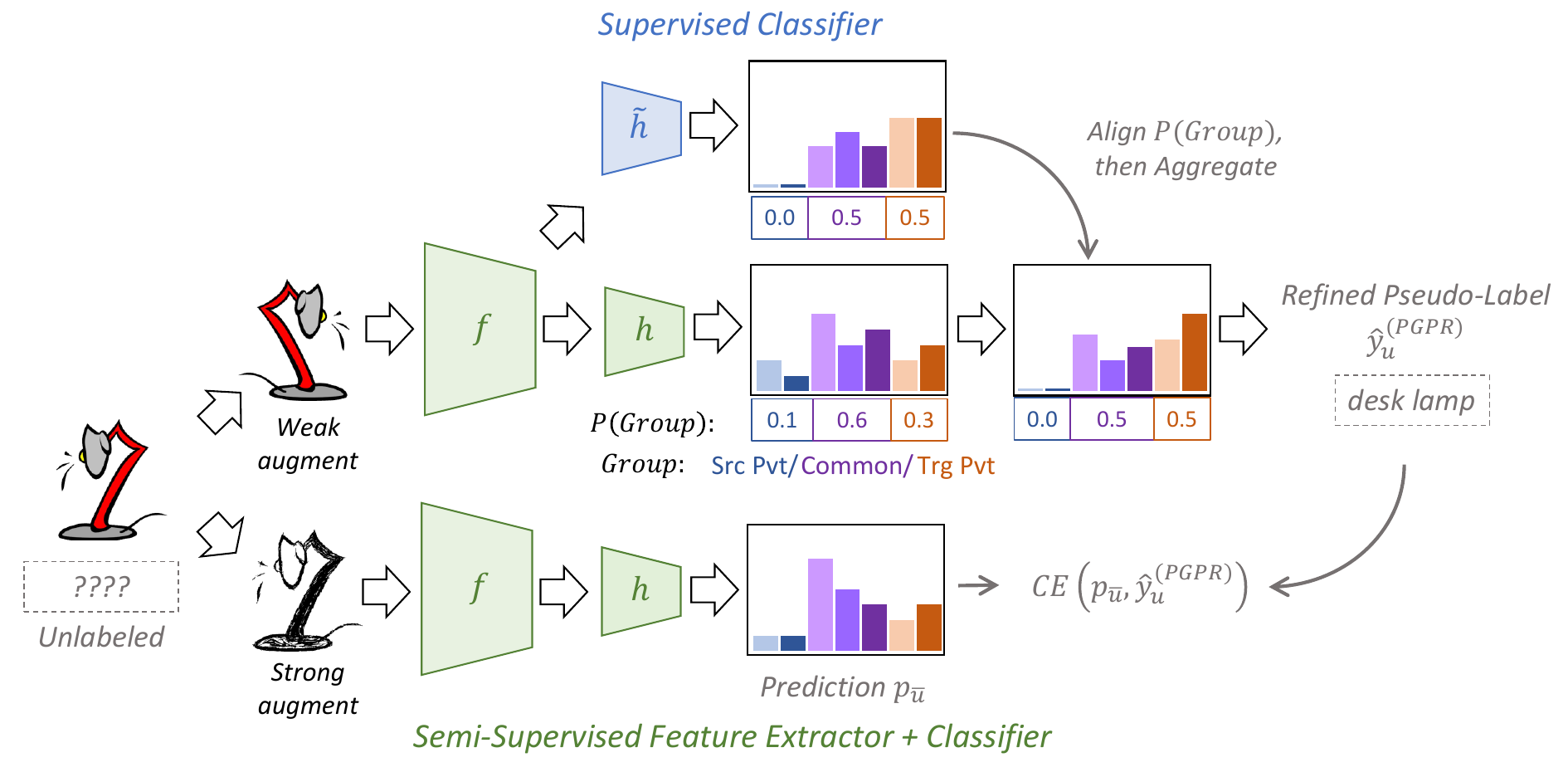}
    
    \vspace{-2mm}
    \caption{Proposed Prior-Guided Pseudo-label Refinement (PGPR) strategy: During adaptation, we add a supervised classification head $\tilde{h}$ trained only on labeled samples to provide prior distribution estimates. We align the per-instance class group distribution output by $h \circ f$ to the estimate by $\tilde{h} \circ f$ and then aggregate the two classifier decisions. The class with the highest resulting probability is taken as the refined pseudo-label for training. The supervised classfier $\tilde{h}$ is discarded at the end of training.}
    \label{fig: proposed_method}
    \vspace{-4mm}
\end{figure}

We denote the input, feature, logit and output space as $\gX, \gZ, \gG$ and $\gY$, respectively.
With a neural network model $h\circ f$, the feature extractor $f$ parameterized by $\Theta$ learns the mapping $f: \gX \rightarrow \gZ$, and classifier $h$ parameterized by $\Psi$ learns the mapping $h: \gZ \rightarrow \gG$. For a logit $g\in \gG$, we obtain the predictive probability $p=\sigma(g)$ using softmax function $\sigma$, and $\hat{y}=\argmax_i(p[i]) \in \gY$ as the predicted label. Let $\ell$ denote the labeled source and target, and $u$ denote the unlabeled target.

\subsection{Supervised vs. Semi-Supervised Classifier}

Due to the abundance of labeled source data, a vanilla model naturally overfits to the source data distribution. This overfitting is especially detrimental to the learning of target private classes in UniSSDA since these classes are not represented at all in the source domain. Consequently, the model focuses learning on classes common to both domains, at the expense of target private classes. We refer to this bias as the `common-class bias'. Naive pseudo-labeling assigns the class with the highest confidence score as the pseudo-label for an unlabeled sample. Using naive pseudo-labels to propagate label information to the unlabeled target samples risks reinforcing pre-existing bias. In the example on open-partial setting in Figure~\ref{fig: classifiers}, the semi-supervised classifier trained with naive pseudo-labels severely underestimates the proportion of target private class samples throughout the training process.
In comparison, by fitting a supervised classification head on top of the same feature extractor, we observe that the supervised classifier is less susceptible to common-class bias and has higher accuracy on target private classes. By refining the pseudo-labels with the supervised classier via our proposed strategy and using them to train the semi-supervised classifier, the refined classifier achieves the highest accuracy. 

\subsection{Prior-Guided Pseudo-Label Refinement}

From Figure~\ref{fig: classifiers}, we observe that the supervised classifier is less susceptible to common-class bias than the naive semi-supervised classifier is.
Motivated by this observation, we propose to refine the predictive probabilities output by $h\circ f$ guided by per-instance priors estimated using a supervised classifier. 
On top of the feature extractor $f$, we add a new linear classification head $\tilde{h}:\mathcal{Z}\rightarrow \mathcal{G}$ parameterized by $\tilde{\Phi}$ learned only on labeled source and target samples. Depending on the domain of the input sample, we apply domain-specific classifier masks to mask logits of classes absent from the domain. The classifier $\tilde{h}$ is trained alongside $h\circ f$ using cross-entropy loss:
\begin{align}
    \tilde{L}_{\ell}(\tilde{\Phi}) &= - y_{\ell} \cdot \log(\tilde{p}_{\ell})
\end{align}
where $\tilde{p}$ is the predictive probability output by $\tilde{h} \circ f$ and $y$ is the ground-truth label. Gradients are blocked in the feature extractor $f$.

The goal of the proposed Prior-Guided Pseudo-label Refinement (PGPR) strategy is to reduce the effect of source-induced bias on target pseudo-labels. On each unlabeled sample, we adjust the 
predictive probability $p_u$ originally estimated by $h \circ f$ using the prior distribution $\tilde{p}_u$ estimated by the supervised classifier $\tilde{h} \circ f$. We apply a group reweighted refinement step followed by a classifier aggregated refinement step. 
Group reweighted refinement aligns the class group distribution. The classes are grouped based on the domain(s) they are present in i.e. source private ($\mathcal{Y}_T^c \cap \mathcal{Y}_S$), target private ($\mathcal{Y}_T \cap \mathcal{Y}_S^c$), and common ($\mathcal{Y}_T \cap \mathcal{Y}_S$) classes. 
We treat $\tilde{p}_{u}$ as the prior to compute the probability of each class group $grp$, and reweigh $p_{u}$ output by $h \circ f$ such that $\sum_{j \in grp}p_{u}^{(reweighted)}[j] = \sum_{j \in grp}\tilde{p}_{u}[j]$, by:
\begin{equation}
    p_{u}^{(reweighted)}[i] = p_{u}[i] \cdot \frac{\sum_{j \in grp}\tilde{p}_{u}[j]}{\sum_{j \in grp}p_{u}[j]} 
\end{equation}
for $i \in grp$. That is, for each class group, the group distribution from $p_{u}^{(reweighted)}$ is instance-wise aligned to that from $\tilde{p}_{u}$.
Class aggregated refinement further adjusts individual class probabilities by aggregating the decisions from the two classifiers to give the final predictive probability:
\begin{equation}
    p_u^{(PGPR)} = \left(p_u^{(reweighted)} + \tilde{p}_u\right)/2.
\end{equation}
The class with the highest probability in $p_u^{(PGPR)}$ is taken as the pseudo-label $\hat{y}_{u}^{(PGPR)}$.

Taken together, group reweighted refinement can be viewed as a strict and coarse-grained regularization on the estimated class group distribution, and classifier aggregated refinement can be viewed as a soft and fine-grained regularization on the estimated class distribution.

\subsection{Training Objectives}

The overall training objective is:
\begin{equation}
    L(\Theta,\Psi) = L_{\ell}(\Theta,\Psi) + \mu(t)L_{u}(\Theta,\Psi) 
\end{equation}
where $L_{\ell}(\Theta,\Psi)$ and $L_{u}(\Theta,\Psi)$ are the respective loss functions on the labeled and unlabeled data, and $\mu(t)=\frac{1}{2} - \frac{1}{2}cos(min(\pi, \frac{\pi t}{T}))$ is a warmup function weighing $L_{u}(\Theta,\Psi)$ at iteration $t$ with total $T$ warmup steps. After training, the supervised classifier $\tilde{h}$ is discarded. Only the model $h\circ f$ is retained for inference.

We apply cross-entropy loss on the labeled data:
\begin{equation} \label{eqn: labeled loss}
    L_{\ell}(\Theta,\Psi) = -y_{\ell} \cdot \log\left(p_{\ell}\right).
\end{equation}
Existing robust training techniques can be incorporated. For instance, AdaMatch \cite{berthelot2021adamatch} interpolates between logit $g_\ell$ and another copy $g'_\ell$ output with augmented batch normalization statistics to compute $p_\ell=\sigma(\lambda \cdot g_\ell + (1-\lambda) \cdot g'_\ell)$, $\lambda\sim Unif(0,1)$, to improve robustness to covariate shift.

The unlabeled data is trained with refined pseudo-labels: 
\begin{align} \label{eqn: unlabeled loss}
    & L_{u}(\Theta,\Psi) \\ \nonumber
    &= - \hat{y}_{u}^{(PGPR)} \cdot \log(p_{\bar{u}}) \cdot \mathds{1}{[\max_j p_{u}^{(PGPR)}[j] \geq c_\tau]} - \\ \nonumber
    & \hspace{0.5cm} \frac{1}{2} \hat{y}_{u}^{(PGPR)} \cdot \log(p_{\bar{u}}) \cdot \mathds{1}{[\max_j p_{u}^{(PGPR)}[j] < c_\tau]}
\end{align}
where pseudo-labels with confidence below threshold $c_\tau = \tau \cdot \E (\max_j p_{\ell}[j])$ have a down-weighted loss. We apply weak and strong augmentations for consistency regularization. In Equation~\ref{eqn: unlabeled loss}, pseudo-labels are estimated on weakly augmented images denoted by $u$, and the loss is applied on strongly augmented images denoted by $\bar{u}$.

\section{Experiments and Results}
\label{sec: experiments and results}

\subsection{Experimental Setups}
\label{sec: experimental setups}

\begin{table*}[tb]

\centering
\begin{adjustbox}{max width=0.75\textwidth}
\begin{tabular}{l*{8}{c}}
\toprule[1pt]\midrule[0.3pt]

\textbf{Adaptation Setting} & \multicolumn{2}{c}{\textbf{Office-Home}} & \multicolumn{2}{c}{\textbf{DomainNet-126}} & \multicolumn{2}{c}{\textbf{DomainNet-345}} & \multicolumn{2}{c}{\textbf{VisDA}} \\ \cmidrule(lr){2-3} \cmidrule(lr){4-5}\cmidrule(lr){6-7}\cmidrule(lr){8-9}
& \textbf{Source} & \textbf{Target} & \textbf{Source} & \textbf{Target} & \textbf{Source} & \textbf{Target} & \textbf{Source} & \textbf{Target} \\ \midrule
Open-set & 1-40 & 1-65 & 1-80 & 1-126 & 1-150 & 1-345 & 1-6 & 1-12\\
Partial-set & 1-65 & 41-65 & 1-126 & 81-126 & 1-345 & 1-150 & 1-12 & 1-6\\
Open-partial & 1-40 & 1-20, 41-65 & 1-80 & 1-40, 81-126 & 1-200 & 1-150, 201-345 & 1-9 & 1-6, 10-12\\
\midrule[0.3pt]\bottomrule[1pt]
\end{tabular}
\end{adjustbox}

\vspace{-2mm}
\caption{Source and target classes in open-set, partial-set and open-partial setting, for Office-Home, DomainNet-126, DomainNet-345 and VisDA. All classes are used in closed-set setting.} \label{tab: adaptation_setting}
\vspace{-4mm}
\end{table*}

\textbf{Methods.} We compare the proposed strategy with the supervised baseline S+T and existing SSDA and UniDA methods modified for UniSSDA.
\textbf{S+T} trains only on labeled source and target samples with cross-entropy loss. 

For existing SSDA methods, we evaluate CDAC, PAC and AdaMatch. \textbf{CDAC}~\cite{li2021cdac} clusters target data and selects highly-confident target samples for pseudo-labeling and consistency regularization. \textbf{PAC}~\cite{mishra2021surprisingly} pre-trains with rotation prediction task and encourages label consistency during adaptation.
\textbf{AdaMatch}~\cite{berthelot2021adamatch} builds on FixMatch~\cite{sohn2020fixmatch} by augmenting data with weak and strong augmentations, applies random logit interpolation for logit alignment and aligns the overall class distribution between domains.

For existing UniDA methods, we evaluate DANCE and UniOT. \textbf{DANCE}~\cite{saito2020dance} uses self-supervision to cluster target samples and uses an entropy threshold to identify samples in common classes for alignment. \textbf{UniOT}~\cite{chang2022uniot} uses optimal transport to cluster target samples around prototypes. It detects samples in common classes for alignment based on statistical information of class assignment estimates, without the need for prefined threshold values.

For all methods except AdaMatch, we apply domain-specific classifier masks to mask logits of classes absent from the domain. We exclude AdaMatch as it interpolates logits across domains during training, but apply the masks during inference to constrain predictions to classes present in the domain. To add target supervision to UniDA methods, we add cross-entropy loss on the labeled target samples to the original training objectives.

\noindent\textbf{Datasets.} We evaluate on 3 popular benchmark datasets for domain adaptation. \textbf{Office-Home}~\cite{Venkateswara2017DeepHN} has 65 categories of everyday objects in 4 domains: Art (A), Clipart (C), Product (P) and Real World (R).
\textbf{DomainNet}~\cite{peng2019moment} has a total of 6 domains and 345 classes. Following \cite{li2021cdac, mishra2021surprisingly}, we evaluate on 4 domains: Clipart (C), Painting (P), Real (R) and Sketch (S). We denote the version with a subset of 126 classes as DomainNet-126 for comparison with SSDA methods \cite{li2021cdac, mishra2021surprisingly}, and denote the full version as DomainNet-345 for comparison with UniDA methods \cite{deng2023uniood}. Each of Office-Home and DomainNet has a total of 12 source-target domain pairs. \textbf{VisDA}~\cite{Peng2017VisDATV} has 12 classes and is used to evaluate synthetic-to-real transfer.

For each domain, we randomly split the samples into 50\% training, 20\% validation, and 30\% testing. Following existing practice \cite{li2021cdac, mishra2021surprisingly}, we assume $k$-shot target annotation for training and validation, with $k$ set to 3 in main experiments. Instead of pre-specifying a single selection of labeled target samples as in existing works \cite{li2021cdac, mishra2021surprisingly}, we randomly draw target samples for labeling from the larger collection of training and validation data in each run, in order to take into account the variability of target annotation.

\noindent\textbf{Adaptation settings.}
We evaluate on the settings listed in Figure~\ref{fig: adaptation_setting}.
For `Closed-set without label distribution shift', we sample the datasets such that all domains share the same sample size and class distribution, to study covariate shift in isolation.
All classes are included in closed-set settings. For the open-set, partial-set and open-partial adaptation settings, we include a subset of classes in the source and/or target domain, as specified in Table~\ref{tab: adaptation_setting}. 

\noindent\textbf{Implementation details.}
For comparison with SSDA methods, we use ResNet-34 backbone plus a linear classifer following \cite{li2021cdac, mishra2021surprisingly}. Models are trained using SGD optimizer with momentum 0.9 and weight decay 0.0005 for 5000 iterations using batch size 24. We set the learning rate 0.001 for the feature extractor and 0.01 for the classifier, and temperature scaling 0.05. 
We standardize the set of image augmentations used to random horizontal flips and crops to $224 \times 224$ for weak augmentations, with the addition of RandAugment for strong augmentations, following \cite{li2021cdac}. 

For comparison with UniDA methods, we follow the training setup in \cite{deng2023uniood} for adapting vision transformer \cite{dosovitskiy2021vit} foundation models: dinov2\_vitl14 in DINOv2 trained with self-supervision \cite{oquab2023dinov2} and ViT-L/14@336px in CLIP trained with image-text pairs \cite{Radford2021clip}. The foundation model encoder is frozen, and the classifier is trained for 10000 iterations using batch size 32, with initial learning rate 0.01 and a cosine scheduler with 50 warmup interations. No data augmentation is applied following \cite{Radford2021clip}.

Experiments are run on NVIDIA container for PyTorch, release 23.02, on NVIDIA GeForce RTX 3090. We set confidence threshold hyperparameter $\tau=0.9$ and warmup step count $T=500$ for our proposed method. Hyperparameters for other methods follow the defaults in \cite{li2021cdac,mishra2021surprisingly,berthelot2021adamatch,deng2023uniood}. Experiments are run over 3 seeds, and we report the average $\rpm$ standard deviation of the target domain accuracy. 

\subsection{Results}
\label{sec: results}

\begin{table}[tb]

\centering
\setlength{\tabcolsep}{4pt}
\begin{adjustbox}{max width=\columnwidth}
\begin{tabular}{l*{6}{c}}
\toprule[1pt]\midrule[0.3pt]

Method & \multicolumn{1}{c}{Covariate Shift} & \multicolumn{4}{c}{Covariate + Label Shift} & \multicolumn{1}{c}{Overall}\\ \cmidrule(lr){2-2} \cmidrule(lr){3-6}
& \multicolumn{1}{c}{Closed-set} & \multicolumn{1}{c}{Closed-set} & \multicolumn{1}{c}{Open-set} & \multicolumn{1}{c}{Partial-set} & \multicolumn{1}{c}{Open-partial} \\ \midrule
\multicolumn{7}{c}{\textbf{Office-Home}} \\ \midrule
S + T & 67.9 \rpm 0.5 & 65.1 \rpm 0.5 & 63.2 \rpm 0.3 & 72.9 \rpm 0.9 & 64.2 \rpm 0.7 & 66.7 \\
CDAC & 69.7 \rpm 0.4 & 67.1 \rpm 1.4 & 60.3 \rpm 0.3 & 68.7 \rpm 1.7 & 56.3 \rpm 1.5 & 64.4 \\
PAC & 67.3 \rpm 0.5 & 65.1 \rpm 0.5 & 60.6 \rpm 0.4 & 69.7 \rpm 1.2 & 61.3 \rpm 0.6 & 64.8 \\
AdaMatch & 69.7 \rpm 0.4 & 66.6 \rpm 0.5 & 63.1 \rpm 0.6 & 74.4 \rpm 0.3 & 64.1 \rpm 0.3 & 67.6 \\ \hdashline
Proposed & \textbf{72.3 \rpm 1.3} & \textbf{69.4 \rpm 0.8} & \textbf{66.9 \rpm 1.2} & \textbf{77.4 \rpm 1.8} & \textbf{67.4 \rpm 1.8} & \textbf{70.7} \\ \midrule

\multicolumn{7}{c}{\textbf{DomainNet-126}} \\ \midrule
S + T & 63.9 \rpm 0.4 & 58.8 \rpm 0.2 & 54.1 \rpm 0.1 & 72.6 \rpm 0.8 & 54.4 \rpm 0.7 & 60.8 \\
CDAC & 69.7 \rpm 0.1 & 65.3 \rpm 0.3 & 52.1 \rpm 0.9 & 75.3 \rpm 0.4 & 43.7 \rpm 1.4 & 61.2 \\
PAC & 69.1 \rpm 0.4 & 64.6 \rpm 0.2 & 51.6 \rpm 0.2 & 77.9 \rpm 0.3 & 51.2 \rpm 0.8 & 62.9 \\
AdaMatch & 66.7 \rpm 0.1 & 61.3 \rpm 0.4 & 53.1 \rpm 0.5 & 76.3 \rpm 0.5 & 53.7 \rpm 1.0 & 62.2 \\ \hdashline
Proposed & \textbf{71.8 \rpm 0.7} & \textbf{67.3 \rpm 0.6} & \textbf{61.2 \rpm 0.8} & \textbf{80.3 \rpm 0.4} & \textbf{62.5 \rpm 1.3} & \textbf{68.6} \\
\midrule[0.3pt]\bottomrule[1pt]
\end{tabular}
\end{adjustbox}

\vspace{-2mm}
\caption{Comparison with SSDA methods: Target accuracy averaged across 12 domain pairs for each dataset, trained with ResNet-34 backbone.} \label{tab: results_average}
\vspace{-4mm}
\end{table}

\begin{table}[tb]

\begin{subtable}{\columnwidth}
\centering
\begin{adjustbox}{max width=0.95\columnwidth}
\begin{tabular}{lP{1.5cm}P{1.5cm}P{1.5cm}P{1.5cm}c}
\toprule[1pt]\midrule[0.3pt]

Method & \multicolumn{1}{c}{Closed-set} & \multicolumn{1}{c}{Open-set} & \multicolumn{1}{c}{Partial-set} & \multicolumn{1}{c}{Open-partial} & \multicolumn{1}{c}{Overall}\\ \midrule
\multicolumn{6}{c}{\textbf{DomainNet-345}} \\ \midrule
S + T & 73.6 \rpm 0.2 & 68.2 \rpm 0.4 & 80.9 \rpm 0.4 & 70.3 \rpm 0.5 & 73.2\\
DANCE & 73.8 \rpm 0.3 & 67.1 \rpm 0.5 & 81.4 \rpm 0.5 & 69.3 \rpm 0.5 & 72.9\\
UniOT & 72.8 \rpm 0.3 & 62.3 \rpm 0.5 & 76.2 \rpm 0.5 & 64.3 \rpm 0.4 & 68.9\\ \hdashline
Proposed & \textbf{74.4 \rpm 0.3} & \textbf{69.0 \rpm 0.4} & \textbf{82.3 \rpm 0.3} & \textbf{71.5 \rpm 0.5} & \textbf{74.3}\\ \midrule

\multicolumn{6}{c}{\textbf{VisDA}} \\ \midrule
S + T & 78.0 \rpm 0.6 & 73.2 \rpm 1.5 & 91.0 \rpm 0.3 & 79.9 \rpm 1.2 & 80.5\\
DANCE & 75.1 \rpm 0.8 & 67.0 \rpm 1.2 & 93.6 \rpm 0.3 & 78.9 \rpm 2.9 & 78.6\\
UniOT & 77.9 \rpm 1.3 & 68.7 \rpm 2.2 & 87.6 \rpm 0.7 & 78.4 \rpm 0.7 & 78.2\\ \hdashline
Proposed & \textbf{79.8 \rpm 0.8} & \textbf{76.5 \rpm 1.5} & \textbf{93.7 \rpm 0.7} & \textbf{82.6 \rpm 0.7} & \textbf{83.2}\\
\midrule[0.3pt]\bottomrule[1pt]
\end{tabular}
\end{adjustbox}
\caption{DINOv2 encoder dinov2\_vitl14}
\end{subtable}

\begin{subtable}{\columnwidth}
\centering
\begin{adjustbox}{max width=0.95\columnwidth}
\begin{tabular}{lP{1.5cm}P{1.5cm}P{1.5cm}P{1.5cm}c}
\toprule[1pt]\midrule[0.3pt]

Method & \multicolumn{1}{c}{Closed-set} & \multicolumn{1}{c}{Open-set} & \multicolumn{1}{c}{Partial-set} & \multicolumn{1}{c}{Open-partial} & \multicolumn{1}{c}{Overall}\\ \midrule
\multicolumn{6}{c}{\textbf{DomainNet-345}} \\ \midrule
S + T & 77.3 \rpm 0.4 & 68.9 \rpm 0.5 & 84.1 \rpm 0.4 & 71.1 \rpm 0.6 & 75.4\\
DANCE & 77.0 \rpm 0.3 & 67.5 \rpm 0.6 & 84.1 \rpm 0.3 & 69.6 \rpm 0.5 & 74.6\\
UniOT & 77.4 \rpm 0.3 & 61.8 \rpm 0.7 & 81.6 \rpm 0.4 & 64.7 \rpm 0.5 & 71.4\\ \hdashline
Proposed & \textbf{77.5 \rpm 0.3} & \textbf{71.1 \rpm 0.7} & \textbf{84.4 \rpm 0.3} & \textbf{73.7 \rpm 0.6} & \textbf{76.7}\\ \midrule

\multicolumn{6}{c}{\textbf{VisDA}} \\ \midrule
S + T & 87.5 \rpm 0.4 & 78.8 \rpm 1.1 & 95.0 \rpm 0.1 & 84.2 \rpm 0.8 & 86.4\\
DANCE & 86.2 \rpm 0.8 & 74.7 \rpm 1.3 & \textbf{96.3 \rpm 0.4} & 82.5 \rpm 1.9 & 84.9\\
UniOT & 87.6 \rpm 1.1 & 79.7 \rpm 0.8 & 92.1 \rpm 1.1 & 84.3 \rpm 0.8 & 85.9\\ \hdashline
Proposed & \textbf{88.0 \rpm 0.1} & \textbf{83.3 \rpm 1.5} & 96.2 \rpm 0.4 & \textbf{84.6 \rpm 0.6} & \textbf{88.0}\\
\midrule[0.3pt]\bottomrule[1pt]
\end{tabular}
\end{adjustbox}
\caption{CLIP encoder ViT-L/14@336px}
\end{subtable}

\vspace{-2mm}
\caption{Comparison with UniDA methods: Target accuracy averaged across 12 domain pairs for DomainNet-345 and synthetic-to-real transfer accuracy for VisDA, in covariate + label shift settings. Training is performed with frozen foundation model encoder and learnable classifier.} \label{tab: results_average_foundation}
\vspace{-4mm}
\end{table}

\subsubsection{Comparison with SSDA Methods}
\label{sec: comparison with ssda}

Table~\ref{tab: results_average} shows the target accuracy averaged across all domain pairs for each dataset. The existing SSDA methods evaluated generally improve over the supervised baseline S+T in the closed-set settings. In the more challenging label space shift settings, CDAC and PAC do not consistently improve performance in the partial-set setting, and degrade performance in the open-set and open-partial settings. Moreover, CDAC learning can be unstable, leading to large performance drops (above 7\%) in open-partial setting. AdaMatch improves performance in the partial-set setting for both datasets, but still marginally degrades performance in the open-set and open-partial setting as a result of common-class bias.

Amongst the SSDA methods evaluated, overall, AdaMatch shows the least performance degradation in non-closed settings. Consequently, we adopt random logit interpolation from AdaMatch for robust training of the labeled samples in Equation~\ref{eqn: labeled loss}. We simulate covariate shifts by passing the input images though the feature extractor $f$ with different batch normalization statistics i.e. statistics from both labeled and unlabeled data and from labeled data alone. We randomly interpolate between the two versions of logits before computing the predictive probability and training with cross-entropy loss.

The proposed method attains the highest overall accuracy, and outperforms the second-best method by 3.1\% and 5.7\% on Office-Home and DomainNet-126, respectively. It consistently outperforms the supervised baseline S+T, and increases accuracy by over 7\% in the challenging open-set and open-partial setting for DomainNet-126. 

\subsubsection{Comparison with UniDA Methods}

Table~\ref{tab: results_average_foundation} shows the target accuracy evaluated on two ViT encoders. The existing UniDA methods evaluated performs similarly or underperforms the supervised baseline S+T in most cases. Existing criteria (e.g. high entropy) to identify target private class samples in UniDA become less suitable in UniSSDA, as a small number of target private class samples are now available for supervision. We observe that performances of all methods are generally higher on the CLIP encoder than on the DINOv2 encoder, as the latter is trained only with self-supervised objectives.
The proposed method attains the highest overall accuracy on both datasets and foundation model encoders. On DomainNet-345, the peformance gain over the second-best method is 1.1\% and 1.3\% with DINOv2 and CLIP, respectively. On VisDA, the performance gain is 2.7\% and 1.6\% with DINOv2 and CLIP, repectively. We note that no special robust training is implemented on the labeled samples to learn robust features as the encoders are frozen. However due to the stronger and more robust feature extraction capability of foundation models, the frozen encoders can more adequately address covariate shift compared to the ResNet-34 in Section~\ref{sec: comparison with ssda}.
\section{Further Analysis}
\label{sec: further analysis}

\noindent \textbf{The proposed method improves private class accuracy without having to sacrifice common class accuracy.}
We study adaptation accuracy on samples in common and target private classes separately in open-set and open-partial settings on DomainNet-345 in Table~\ref{tab: results_common_private_foundation}. We observe that the UniDA method UniOT achieves the highest common class accuracy in 5 out of 8 cases, but at the expense of private class accuracy. Comparing methods with similar common class accuracy, the proposed method improves private class accuracy in most cases. We include comparisons with SSDA methods in the Appendix.

\noindent \textbf{The proposed method is effective in scenarios where not all target classes are labeled.} We study these scenarios on DomainNet-345 in Table~\ref{tab: results_unknown}. In `Unlabeled Private', `Unlabeled Common' and `Unlabeled Mixed', we set 45 target private classes (class 301-345), 50 common classes (class 101-150) and a mixture of 95 classes (class 301-345, 101-150) as entirely unlabeled, respectively. Following UniDA practice, we treat unlabeled private classes as a single `unknown' class. For all non-UniDA methods, we adopt the one-vs-all classifier from OVANet~\cite{saito2021ovanet} to detect `unknown' samples. Pre-training method PAC is excluded since the encoder is frozen. We include recent methods for long-tailed learning (DRw~\cite{peng2023drw}), SSDA (ProML~\cite{huang2023proml}) and UniDA (OVANet~\cite{saito2021ovanet}). Our method attains the best performance in these challenging scenarios.
\begin{table}[tb]

\centering
\begin{adjustbox}{max width=0.95\columnwidth}
\begin{tabular}{l*{2}{c}*{2}{c}}
\toprule[1pt]\midrule[0.3pt]

Method & Open-set & Open-partial & Open-set & Open-partial \\
& (Common / Pvt) & (Common / Pvt) & (Common / Pvt) & (Common / Pvt) \\ \midrule
& \multicolumn{2}{c}{\textbf{DomainNet-345 (CLIP)}} & \multicolumn{2}{c}{\textbf{VisDA (CLIP)}} \\ \midrule
S + T       & 81.5 / 60.0 & 81.9 / 61.4 & 92.7 / 65.0 & 92.7 / \textbf{67.2}\\
DANCE       & 81.1 / 57.8 & 81.2 / 59.0 & 91.0 / 58.4 & 92.7 / 62.2 \\
UniOT       & \textbf{83.0} / 46.7 & \textbf{82.3} / 48.7 & \textbf{94.9} / 64.5 & 93.5 / 66.1 \\ \hdashline
Proposed    & 81.2 / \textbf{64.0} & \textbf{82.3} / \textbf{66.0} & 94.1 / \textbf{72.4} & \textbf{93.9} / 65.8 \\
\midrule[0.3pt]\bottomrule[1pt]
\end{tabular}
\end{adjustbox}

\vspace{-2mm}
\caption{Average target accuracy on common and target private (Ptv) classes.} \label{tab: results_common_private_foundation_CLIP}
\vspace{-4mm}
\end{table}
\begin{table}[tb]

\centering
\begin{adjustbox}{max width=0.9\columnwidth}
\begin{tabular}{l*{4}{c}}
\toprule[1pt]\midrule[0.3pt]

Method & \multicolumn{1}{c}{Open-partial} & \multicolumn{1}{c}{Unl. Private} & \multicolumn{1}{c}{Unl. Common} & \multicolumn{1}{c}{Unl. Mixed}  \\ \midrule

S + T       & 71.1 \rpm 0.6 & 65.5 \rpm 0.7 & 70.5 \rpm 0.6 & 64.1 \rpm 0.8 \\
DRw         & 69.2 \rpm 0.7 & 65.0 \rpm 0.8 & 62.1 \rpm 0.5 & 59.0 \rpm 0.9 \\
CDAC        & 69.3 \rpm 0.6 & 64.6 \rpm 0.6 & 67.3 \rpm 0.5 & 63.0 \rpm 0.6 \\
AdaMatch    & 68.6 \rpm 0.6 & 65.1 \rpm 0.7 & 68.1 \rpm 0.6 & 64.1 \rpm 0.7 \\
ProML       & 71.1 \rpm 0.6 & 66.1 \rpm 0.7 & 70.5 \rpm 0.5 & 64.9 \rpm 0.8 \\
DANCE       & 69.6 \rpm 0.5 & 66.0 \rpm 0.8 & 71.5 \rpm 0.7 & 64.6 \rpm 0.8 \\
UniOT       & 64.7 \rpm 0.5 & 58.8 \rpm 0.4 & 64.9 \rpm 0.6 & 58.9 \rpm 0.5 \\
OVANet      & 71.2 \rpm 0.6 & 66.1 \rpm 0.7 & 70.5 \rpm 0.5 & 64.9 \rpm 0.8  \\ \hdashline
Proposed    & \textbf{73.7 \rpm 0.6} & \textbf{67.1 \rpm 0.6} & \textbf{72.5 \rpm 0.5} & \textbf{66.4 \rpm 0.6}  \\
\midrule[0.3pt]\bottomrule[1pt]
\end{tabular}
\end{adjustbox}

\vspace{-2mm}
\caption{Comparison with long-tailed SSL, SSDA and UniDA methods: Accuracy averaged across 12 domain pairs for DomainNet-345 with CLIP encoder. In Unlabeled (Unl.) Private / Common / Mixed scenarios, several target private / common / mixture of private and common classes are entirely unlabeled.} \label{tab: results_unknown}
\vspace{-4mm}
\end{table}

\noindent \textbf{Each pseudo-label refinement step is effective in improving pseudo-label quality.}
We perform the ablation study on DomainNet-126 $C\rightarrow P$ across all UniSSDA settings. Note that the group reweighted refinement step does not affect adaptation in closed-set settings since all classes belong to the common class group. From Table~\ref{tab: ablation}, overall, each refinement step helps to improve target accuracy.
\begin{table*}[tb]

\centering
\begin{adjustbox}{max width=0.8\textwidth}
\begin{tabular}{P{3cm}P{3cm}*{6}{c}}
\toprule[1pt]\midrule[0.3pt]

Group reweighted & Classifier aggregated & \multicolumn{1}{c}{Covariate Shift} & \multicolumn{4}{c}{Covariate + Label Shift} & \multicolumn{1}{c}{Overall}\\ \cmidrule(lr){3-3} \cmidrule(lr){4-7}
\multicolumn{2}{c}{} & \multicolumn{1}{c}{Closed-set} & \multicolumn{1}{c}{Closed-set} & \multicolumn{1}{c}{Open-set} & \multicolumn{1}{c}{Partial-set} & \multicolumn{1}{c}{Open-partial} \\ \midrule
\xmark & \xmark & 66.1 & 61.6 & 50.4 & \textbf{78.7} & 51.5 & 61.7\\
\cmark & \xmark & 66.1 & 61.6 & 56.8 & \textbf{78.7} & 59.2 & 64.5\\
\cmark & \cmark & \textbf{66.9} & \textbf{63.7} & \textbf{57.4} & 78.5 & \textbf{60.8} & \textbf{65.5}\\
\midrule[0.3pt]\bottomrule[1pt]
\end{tabular}
\end{adjustbox}

\vspace{-2mm}
\caption{Ablation study on effectivess of pseudo-label refinement steps, evaluated on DomainNet-126 $C\rightarrow P$.} \label{tab: ablation}
\vspace{-2mm}
\end{table*}

\noindent \textbf{The refined pseudo-labels can be readily incorporated into existing SSDA methods to expand their adaptation capabilities to non-closed-set settings.}
In existing SSDA methods, we add a supervised classification head on top of the feature extractor to estimate prior distributions for pseduo-label refinement.
We replace the original pseudo-labels used with our refined pseudo-labels while retaining all other components of the algorithms. From Table~\ref{tab: results_average_add_refinement}, we see up to 3.4\%, 3.9\% and 7.8\% improvement in open-set, partial-set and open-partial setting, respectively.
\begin{table}[tb]

\centering
\begin{adjustbox}{max width=0.8\columnwidth}
\begin{tabular}{l*{3}{l}}
\toprule[1pt]\midrule[0.3pt]

Method & \multicolumn{1}{c}{Open-set} & \multicolumn{1}{c}{Partial-set} & \multicolumn{1}{c}{Open-partial} \\ \midrule

CDAC & 50.0 & 74.8 & 47.2\\
\quad + PGPR & 49.9 $\red{(\downarrow 0.1)}$ & 74.9 $\green{(\uparrow 0.1)}$ & 55.0 $\green{(\uparrow 7.8)}$ \\ \hdashline
PAC & 46.8 & 75.2 & 46.9 \\
\quad + PGPR & 49.2 $\green{(\uparrow 2.4)}$ & 77.7 $\green{(\uparrow 2.5)}$ & 53.5 $\green{(\uparrow 6.6)}$\\ \hdashline
AdaMatch & 50.2 & 73.9 & 51.9 \\
\quad + PGPR & 53.6 $\green{(\uparrow 3.4)}$ & 77.8 $\green{(\uparrow 3.9)}$ & 58.5 $\green{(\uparrow 6.6)}$ \\
\midrule[0.3pt]\bottomrule[1pt]
\end{tabular}
\end{adjustbox}

\vspace{-2mm}
\caption{Proposed prior-guided pseudo-label refinement can be incorporated into existing SSDA methods to expand their capabilities to non-closed-set settings. Values in \green{green}/\red{red} denote \green{increase}/\red{decrease} over performance of the original method, evaluated on DomainNet-126 $C \rightarrow P$.} \label{tab: results_average_add_refinement}
\vspace{-4mm}
\end{table}

\begin{figure*}[htb]
     \centering
     \begin{subfigure}[b]{0.25\textwidth}
         \centering
         \includegraphics[width=\textwidth]{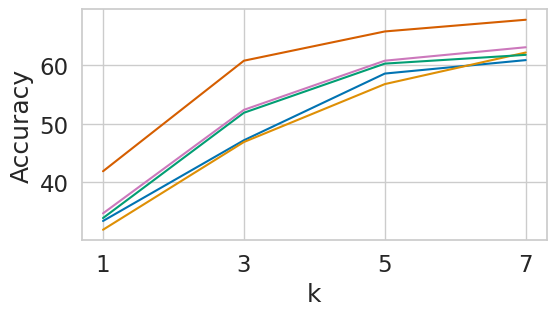}
         \caption{Vary annotation}
         \label{fig: vary_k}
     \end{subfigure}
     \begin{subfigure}[b]{0.25\textwidth}
         \centering
         \includegraphics[width=\textwidth]{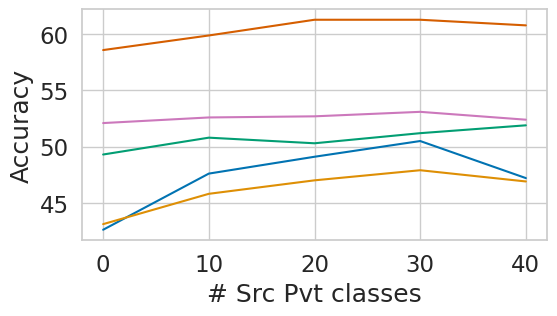}
         \caption{Vary `partialness'}
         \label{fig: inc_src_pvt_classes}
     \end{subfigure}
     \begin{subfigure}[b]{0.35\textwidth}
         \centering
         \includegraphics[width=\textwidth]{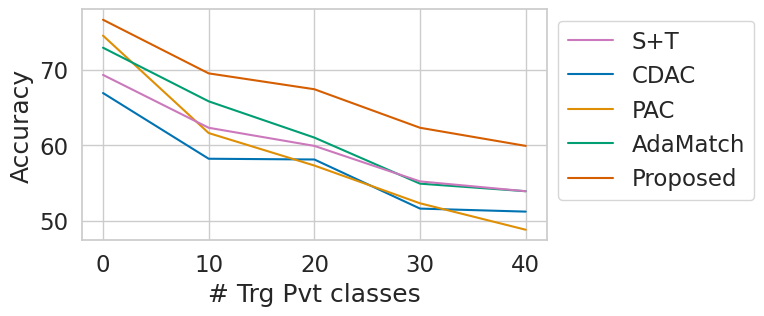}
         \caption{Vary `openness'}
         \label{fig: inc_trg_pvt_classes}
     \end{subfigure}
     
    \vspace{-2mm}
    \caption{Further experiments on DomainNet-126 $C\rightarrow P$ in open-partial setting. (a) plots target accuracy with different $k$ for $k$-shot target annotation. (b) and (c) plot target accuracy under varying degrees of `partialness' and `openness', respectively.}
    \label{fig: further_analysis}
    \vspace{-4mm}
\end{figure*}

\noindent \textbf{The proposed method is effective with varying amounts of target annotation.} We vary $k\in\{1,3,5,7\}$ for $k$-shot target annotation in Figure~\ref{fig: vary_k}. We focus on the most challenging open-partial setting. On DomainNet-126 $C\rightarrow P$, the proposed method outperforms SSDA methods by more than 4\% on all values of $k$.

\noindent \textbf{The proposed method is effective under varying degrees of `partialness' and `openness'.} We experiment with different number of source or target private classes on DomainNet-126 $C\rightarrow P$ to vary the degree of `partialness' or `openness' in the open-partial setting. In Figure~\ref{fig: inc_src_pvt_classes}, we start with 40 common classes (class 1-40), 46 target private classes (class 81-126) and no source private class, and increase the number of source private classes from 0 to 40 by increments of 10 (class 41-80). The proposed method outperforms SSDA methods by more than 6\%. We observe that target accuracy tends to slightly increase when source private classes are initially added till 30 classes and then decrease. Source private class samples may help feature learning, but since these classes are irrelevant to the target domain, an abundance of these samples may distract the model from learning target-relevant features.

In Figure~\ref{fig: inc_trg_pvt_classes}, we start with 40 common classes (class 1-40), 40 source private classes (class 41-80) and no target private class, and increase the number of target private classes from 0 to 40 by increments of 10 (class 81-120). Target accuracy expectedly decreases as the number of target classes increases. The advantage of the proposed method is more evident under higher degree of `openness'. The proposed method outperforms all other methods by 2.1\% at no target private classes, and 6\% at 40 target private classes.

\begin{figure}[tb]
    \centering
    \includegraphics[width=\columnwidth]{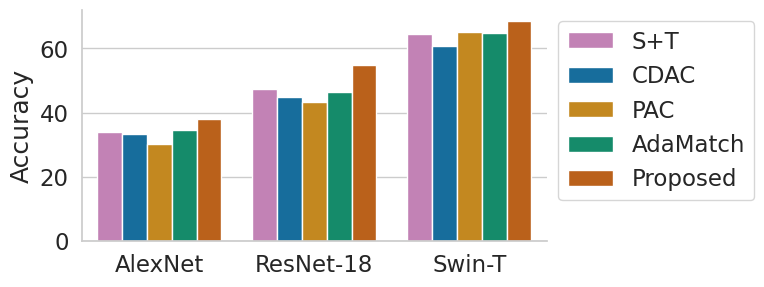}
     
    \vspace{-2mm}
    \caption{Further comparison with existing SSDA methods using different model backbones, evaluated on DomainNet-126 $C\rightarrow P$ in open-partial setting.}
    \label{fig: vary_backbone}
    \vspace{-4mm}
\end{figure}

\noindent \textbf{The proposed method can be applied effectively on different model backbones.} We additionally compare against SSDA methods on other backbone architectures including AlexNet, ResNet-18 and Swin-T (Tiny version of Swin Transformer \cite{liu2021Swin}). From Figure~\ref{fig: vary_backbone}, in open-partial setting on DomainNet-126 $C\rightarrow P$, the proposed method outperforms the second-best method by 3.9\%, 7.5\% and 3.2\% on AlexNet, ResNet-18 and Swin-T, respectively.

\section{Conclusion}
\label{sec: conclusion}

This work introduces Universal SSDA, a generalized semi-supervised domain adaptation problem covering diverse and practical types of covariate and label shifts. We find that existing SSDA and UniDA methods are susceptible to common-class bias and do not consistently perform well in UniSSDA settings. We propose a new prior-guided pseudo-label refinement strategy to address the reinforcement of common-class bias during label propagation. The proposed strategy is simple to implement and effective, as demonstrated through evaluations on multiple datasets and models. We propose the approach as a baseline for further research and application in this area.

\section*{Acknowledgments}
This research is supported by the Agency for Science, Technology and Research (A*STAR) under its AME Programmatic Funds (Grant No. A20H6b0151).

\clearpage

{
    \small
    \bibliographystyle{ieeenat_fullname}
    \bibliography{references}
}

\clearpage
\appendix
\section*{Appendix}

\section{Experiment Details}

We provide additional details on datasets used, experimental results and analysis.

\subsection{Datasets}

We pre-process the datasets according to the adaptation setting. For Office-Home and DomainNet-126 in the closed-set setting with no label distribution shift, we construct the dataset such that sample size per class is the same across domains. We sample the dataset by setting the class size as the minimum size of that class across all domains. For each domain, we randomly split the samples into 50\% training, 20\% validation and 30\% testing.
For the label distribution shift setting, we maintain the sample size of each split, but sample the datasets according to the original class distribution for each domain. For the label space shift settings, we remove classes from the source and/or target domains according to Table 1 in main manuscript. For DomainNet-345 and VisDA in the label shift settings, we directly use the data in \cite{deng2023uniood} and split it into training, validation and testing sets.

\subsection{Results}

\subsubsection{Method Effectiveness}

We report detailed results of adaptation accuracy for each of the 12 source-target domain pairs in Office-Home and DomainNet-126 in Table~\ref{tab: results_full_officehome} and \ref{tab: results_full_domainnet}, DomainNet-345 adapted with DINOv2 encoder in Table~\ref{tab: results_full_domainnet345_dino}, and DomainNet-345 adapted with CLIP encoder in Table~\ref{tab: results_full_domainnet345_clip}. With our prior-guided pseudo-labeling refinement strategy, the proposed method achieves the best performance on the vast majority of domain pairs across the adaptation settings tested. On the remaining domain pairs, it achieves the second-best performance in most cases.

\subsubsection{Private and Common Class Accuracy}

The proposed method improves private class accuracy without having to sacrifice common class accuracy.
We study adaptation accuracy on samples in common and target private classes separately in open-set and open-partial settings. In Table~\ref{tab: results_common_private}, while SSDA methods CDAC, PAC and AdaMatch can achieve performance gains over S+T on common classes, they suffer drastic performance degradation on private classes. For AdaMatch, accuracy on private classes can be lower than that on common classes by approximately 40\% in open-set setting and 30\% in open-partial setting. With the proposed method, although the performance gap between the two class groups still exists, it generally improves private class accuracy without sacrifing common class accuracy. In Table~\ref{tab: results_common_private_foundation}, we observe that the UniDA method UniOT achieves the highest common class accuracy in 5 out of 8 cases, but at the expense of private class accuracy. Comparing methods with similar common class accuracy, the proposed method improves private class accuracy in most cases.
\begin{table}[tb]

\centering
\setlength{\tabcolsep}{4pt}
\begin{adjustbox}{max width=0.95\columnwidth}
\begin{tabular}{l*{4}{c}}
\toprule[1pt]\midrule[0.3pt]

Method & Open-set & Open-partial & Open-set & Open-partial\\
& (Common / Pvt) & (Common / Pvt) & (Common / Pvt) & (Common / Pvt) \\ \midrule
& \multicolumn{2}{c}{\textbf{Office-Home}} & \multicolumn{2}{c}{\textbf{DomainNet-126}} \\ \midrule
S + T & 72.6 / 43.7 & 75.0 / 52.6 & 66.2 / 40.4 & 69.8 / 47.0\\
CDAC & 74.3 / 31.3 & 67.4 / 44.6 & 69.9 / 32.0 & 61.8 / 35.1 \\
PAC & 75.2 / 30.4 & \textbf{80.6} / 40.8 & 69.4 / 31.4 & 73.2 / 40.8 \\
AdaMatch & 76.3 / 35.5 & \textbf{80.6} / 46.5 & 70.4 / 33.5 & \textbf{74.8} / 43.7 \\ \hdashline
Proposed & \textbf{76.7} / \textbf{46.7} & 79.2 / \textbf{54.8} & \textbf{72.4} / \textbf{48.5} & 74.7 / \textbf{56.7}\\
\midrule[0.3pt]\bottomrule[1pt]
\end{tabular}
\end{adjustbox}

\vspace{-2mm}
\caption{Average target accuracy on common and target private (Ptv) classes, trained with ResNet-34 backbone.} \label{tab: results_common_private}
\vspace{-2mm}
\end{table}

\begin{table}[tb]

\centering
\begin{adjustbox}{max width=0.95\columnwidth}
\begin{tabular}{l*{2}{c}*{2}{c}}
\toprule[1pt]\midrule[0.3pt]

Method & Open-set & Open-partial & Open-set & Open-partial \\
& (Common / Pvt) & (Common / Pvt) & (Common / Pvt) & (Common / Pvt) \\ \midrule
& \multicolumn{2}{c}{\textbf{DomainNet-345 (DINOv2)}} & \multicolumn{2}{c}{\textbf{VisDA (DINOv2)}} \\ \midrule
S + T       & 77.1 / 62.0 & \textbf{77.8} / 63.8 & 85.6 / 60.8 & 87.8 / \textbf{64.2}\\
DANCE       & 77.2 / 60.0 & 77.6 / 62.1 & 80.9 / 53.1 & 87.1 / 62.5 \\
UniOT       & \textbf{78.6} / 50.8 & 77.4 / 52.8 & \textbf{89.9} / 47.6 & 88.7 / 57.8 \\ \hdashline
Proposed    & 76.4 / \textbf{63.8} & \textbf{77.8} / \textbf{66.1} & 87.9 / \textbf{65.1} & \textbf{92.0} / 63.9 \\ \midrule
& \multicolumn{2}{c}{\textbf{DomainNet-345 (CLIP)}} & \multicolumn{2}{c}{\textbf{VisDA (CLIP)}} \\ \midrule
S + T       & 81.5 / 60.0 & 81.9 / 61.4 & 92.7 / 65.0 & 92.7 / \textbf{67.2}\\
DANCE       & 81.1 / 57.8 & 81.2 / 59.0 & 91.0 / 58.4 & 92.7 / 62.2 \\
UniOT       & \textbf{83.0} / 46.7 & \textbf{82.3} / 48.7 & \textbf{94.9} / 64.5 & 93.5 / 66.1 \\ \hdashline
Proposed    & 81.2 / \textbf{64.0} & \textbf{82.3} / \textbf{66.0} & 94.1 / \textbf{72.4} & \textbf{93.9} / 65.8 \\
\midrule[0.3pt]\bottomrule[1pt]
\end{tabular}
\end{adjustbox}

\vspace{-2mm}
\caption{Average target accuracy on common and target private (Ptv) classes, trained with frozen foundation model encoder and learnable classifier.} \label{tab: results_common_private_foundation}
\vspace{-2mm}
\end{table}

\subsubsection{Transductive Performance}

In Table~\ref{tab: results_average_transductive}, we provide results on the transductive performance of the proposed method on Office-Home and DomainNet-126. Classification accuracy is measured on the unlabeled target samples used in training instead of the test set. The proposed method achieves the best overall transductive performance on both datasets.
Interestingly, transductive accuracy is lower than inductive accuracy in some cases, e.g. all methods in Office-Home, as the models overfit wrong pseudo-labels to specific training examples.
\begin{table}[tb]

\centering
\begin{adjustbox}{max width=\columnwidth}
\begin{tabular}{l*{6}{c}}
\toprule[1pt]\midrule[0.3pt]

Method & \multicolumn{1}{c}{Covariate Shift} & \multicolumn{4}{c}{Covariate + Label Shift} & \multicolumn{1}{c}{Overall}\\ \cmidrule(lr){2-2} \cmidrule(lr){3-6}
& \multicolumn{1}{c}{Closed-set} & \multicolumn{1}{c}{Closed-set} & \multicolumn{1}{c}{Open-set} & \multicolumn{1}{c}{Partial-set} & \multicolumn{1}{c}{Open-partial} \\ \midrule
\multicolumn{7}{c}{\textbf{Office-Home}} \\ \midrule
CDAC & 68.8 \rpm 0.5 & 67.1 \rpm 1.0 & 59.5 \rpm 0.3 & 68.2 \rpm 1.3 & 52.8 \rpm 1.5 & 63.3 \\
PAC & 67.4 \rpm 0.6 & 65.7 \rpm 0.6 & 59.9 \rpm 0.6 & 70.0 \rpm 1.2 & 60.6 \rpm 0.7 & 64.7 \\
AdaMatch & 67.3 \rpm 1.0 & 66.2 \rpm 0.6 & 62.2 \rpm 0.3 & 72.4 \rpm 0.7 & 63.7 \rpm 0.5 & 66.4 \\ \hdashline
Proposed & \textbf{71.6 \rpm 1.0} & \textbf{70.1 \rpm 1.0} & \textbf{65.0 \rpm 1.0} & \textbf{76.5 \rpm 2.0} & \textbf{66.1 \rpm 1.4} & \textbf{69.9}\\ \midrule

\multicolumn{7}{c}{\textbf{DomainNet-126}} \\ \midrule
CDAC & 70.8 \rpm 0.2 & 66.1 \rpm 0.1 & 52.8 \rpm 0.9 & 73.4 \rpm 0.4 & 44.3 \rpm 1.3 & 61.5 \\
PAC & 69.7 \rpm 0.3 & 64.9 \rpm 0.4 & 52.0 \rpm 0.2 & 78.0 \rpm 0.1 & 51.6 \rpm 0.3 & 63.2 \\
AdaMatch & 66.5 \rpm 0.3 & 61.2 \rpm 0.2 & 53.0 \rpm 0.6 & 75.4 \rpm 0.9 & 53.7 \rpm 0.7 & 61.9\\ \hdashline
Proposed & \textbf{72.3 \rpm 0.5} & \textbf{67.5 \rpm 0.5} & \textbf{60.1 \rpm 0.7} & \textbf{80.1 \rpm 1.0} & \textbf{61.0 \rpm 1.0} & \textbf{68.2}\\
\midrule[0.3pt]\bottomrule[1pt]
\end{tabular}
\end{adjustbox}

\vspace{-2mm}
\caption{Transductive target accuracy averaged across 12 domain pairs for each dataset, trained with ResNet-34 backbone. Note S+T is excluded as it does not train with unlabeled target data. \label{tab: results_average_transductive}}
\vspace{-2mm}
\end{table}

\begin{table*}[tbh]

\begin{subtable}{\textwidth}
\centering
\begin{adjustbox}{max width=\columnwidth}
\begin{tabular}{c|*{5}{c}|*{5}{c}}
\toprule[1pt]\midrule[0.3pt]

\textbf{S $\rightarrow$ T}  & \multicolumn{5}{c|}{\textbf{Closed-set w/o Label Distribution Shift}}           
                            & \multicolumn{5}{c}{\textbf{Closed-set w/ Label Distribution Shift}} \\ \cmidrule(lr){2-6} \cmidrule(lr){7-11}
& S + T & CDAC & PAC & AdaMatch & \makecell{Proposed} & S + T & CDAC & PAC & AdaMatch & \makecell{Proposed}\\ \midrule
A $\rightarrow$ C & 60.3 \rpm 0.7 & 62.2 \rpm 1.0 & 64.0 \rpm 2.4 & 62.8 \rpm 1.0 & \textbf{66.7 \rpm 1.8} & 55.6 \rpm 1.2 & \textbf{60.8 \rpm 1.1} & 57.9 \rpm 1.0 & 59.4 \rpm 1.1 & 60.3 \rpm 0.6\\
A $\rightarrow$ P & 73.2 \rpm 0.8 & 76.3 \rpm 1.6 & 75.0 \rpm 0.5 & 76.3 \rpm 0.5 & \textbf{77.7 \rpm 1.9} & 72.9 \rpm 1.1 & 74.1 \rpm 0.2 & 71.9 \rpm 0.8 & 75.4 \rpm 0.6 & \textbf{77.0 \rpm 1.3}\\
A $\rightarrow$ R & 74.4 \rpm 1.2 & 75.1 \rpm 1.0 & 73.0 \rpm 0.5 & 75.2 \rpm 0.3 & \textbf{75.9 \rpm 1.0} & 74.9 \rpm 0.7 & 74.3 \rpm 0.6 & 71.3 \rpm 2.1 & 74.4 \rpm 0.7 & \textbf{75.4 \rpm 0.5}\\
C $\rightarrow$ A & 58.4 \rpm 0.7 & 62.0 \rpm 0.7 & 58.1 \rpm 1.4 & 61.4 \rpm 1.9 & \textbf{64.1 \rpm 2.1} & 56.5 \rpm 0.5 & 59.4 \rpm 3.2 & 54.7 \rpm 1.8 & 59.0 \rpm 0.9 & \textbf{62.0 \rpm 0.2}\\
C $\rightarrow$ P & 71.6 \rpm 2.1 & 72.6 \rpm 1.8 & 70.4 \rpm 1.6 & 72.9 \rpm 1.1 & \textbf{77.5 \rpm 2.4} & 70.7 \rpm 1.3 & 73.9 \rpm 1.5 & 71.3 \rpm 0.5 & 71.6 \rpm 1.3 & \textbf{76.4 \rpm 0.8}\\
C $\rightarrow$ R & 72.7 \rpm 1.8 & 73.8 \rpm 0.9 & 67.2 \rpm 1.3 & 71.5 \rpm 1.8 & \textbf{74.7 \rpm 0.4} & 69.5 \rpm 0.4 & 70.7 \rpm 1.0 & 64.9 \rpm 1.9 & 70.6 \rpm 1.1 & \textbf{72.1 \rpm 0.3}\\
P $\rightarrow$ A & 61.9 \rpm 1.6 & 63.3 \rpm 1.9 & 57.2 \rpm 1.4 & 63.3 \rpm 1.4 & \textbf{67.9 \rpm 1.2} & 56.8 \rpm 0.7 & 56.7 \rpm 4.1 & 58.0 \rpm 0.2 & 57.7 \rpm 1.1 & \textbf{62.0 \rpm 0.8}\\
P $\rightarrow$ C & 57.6 \rpm 1.6 & 61.3 \rpm 0.3 & 61.9 \rpm 2.2 & 62.2 \rpm 1.1 & \textbf{65.8 \rpm 2.1} & 53.6 \rpm 0.4 & 56.4 \rpm 1.6 & 58.0 \rpm 0.5 & 56.3 \rpm 2.8 & \textbf{60.2 \rpm 1.2}\\
P $\rightarrow$ R & \textbf{77.4 \rpm 0.3} & 77.3 \rpm 0.6 & 73.9 \rpm 0.6 & 77.3 \rpm 0.2 & \textbf{77.4 \rpm 1.0} & 75.9 \rpm 2.3 & 75.0 \rpm 1.8 & 71.9 \rpm 1.1 & 75.7 \rpm 0.1 & \textbf{76.1 \rpm 0.8}\\
R $\rightarrow$ A & 68.2 \rpm 0.7 & 69.3 \rpm 1.2 & 65.7 \rpm 0.5 & 69.5 \rpm 0.1 & \textbf{71.1 \rpm 0.4} & 63.0 \rpm 0.8 & 64.0 \rpm 3.7 & 64.1 \rpm 0.4 & 62.9 \rpm 2.2 & \textbf{68.6 \rpm 1.4}\\
R $\rightarrow$ C & 59.9 \rpm 0.7 & 63.3 \rpm 1.2 & 63.5 \rpm 0.9 & 64.4 \rpm 0.5 & \textbf{66.6 \rpm 0.6} & 56.6 \rpm 0.8 & 61.5 \rpm 2.0 & 60.7 \rpm 1.7 & 59.2 \rpm 1.1 & \textbf{63.5 \rpm 0.6}\\
R $\rightarrow$ P & 79.7 \rpm 1.4 & 79.5 \rpm 1.8 & 78.1 \rpm 1.1 & 79.1 \rpm 0.9 & \textbf{82.0 \rpm 1.3} & 75.9 \rpm 1.2 & 78.2 \rpm 1.7 & 76.9 \rpm 0.8 & 76.9 \rpm 0.5 & \textbf{79.4 \rpm 0.8}\\ \midrule
Average & 67.9 & 69.7 & 67.3 & 69.7 & \textbf{72.3} & 65.1 & 67.1 & 65.1 & 66.6 & \textbf{69.4}\\
\midrule[0.3pt]\bottomrule[1pt]
\end{tabular}
\end{adjustbox}
\caption{Closed set and class distribution shift settings \label{tab: results_full_officehome_closed_cds}}
\end{subtable}

\begin{subtable}{\textwidth}
\centering
\begin{adjustbox}{max width=\columnwidth}
\begin{tabular}{c|*{5}{c}|*{5}{c}}
\toprule[1pt]\midrule[0.3pt]

\textbf{S $\rightarrow$ T}  & \multicolumn{5}{c|}{\textbf{Open-set}}           
                            & \multicolumn{5}{c}{\textbf{Partial-set}} \\ \cmidrule(lr){2-6} \cmidrule(lr){7-11}
& S + T & CDAC & PAC & AdaMatch & \makecell{Proposed} & S + T & CDAC & PAC & AdaMatch & \makecell{Proposed}\\ \midrule
A $\rightarrow$ C & 54.6 \rpm 1.9 & 55.3 \rpm 2.0 & 57.3 \rpm 1.2 & 57.8 \rpm 1.6 & \textbf{60.0 \rpm 0.7} & 61.7 \rpm 3.1 & 56.9 \rpm 2.4 & 60.0 \rpm 2.5 & 69.5 \rpm 3.0 & \textbf{71.5 \rpm 1.5}\\
A $\rightarrow$ P & 70.1 \rpm 2.1 & 68.1 \rpm 2.1 & 68.2 \rpm 2.0 & 72.2 \rpm 1.8 & \textbf{76.4 \rpm 0.8} & 81.4 \rpm 2.6 & 79.4 \rpm 4.3 & 79.4 \rpm 1.6 & 81.9 \rpm 1.2 & \textbf{84.3 \rpm 1.0}\\
A $\rightarrow$ R & 70.0 \rpm 0.6 & 64.1 \rpm 2.0 & 64.2 \rpm 1.9 & 67.6 \rpm 1.4 & \textbf{71.0 \rpm 0.9} & 78.4 \rpm 1.9 & 74.6 \rpm 1.6 & 76.7 \rpm 2.8 & 80.2 \rpm 2.4 & \textbf{82.6 \rpm 1.4}\\
C $\rightarrow$ A & 54.5 \rpm 1.0 & 52.4 \rpm 1.4 & 49.5 \rpm 2.9 & 53.4 \rpm 1.9 & \textbf{59.3 \rpm 2.3} & 68.2 \rpm 1.0 & 63.3 \rpm 3.0 & 62.2 \rpm 1.5 & 69.6 \rpm 1.9 & \textbf{70.2 \rpm 2.0}\\
C $\rightarrow$ P & 69.8 \rpm 0.7 & 67.1 \rpm 0.6 & 66.6 \rpm 1.0 & 70.4 \rpm 2.3 & \textbf{75.0 \rpm 1.9} & 80.2 \rpm 1.9 & 75.8 \rpm 1.2 & 76.1 \rpm 1.2 & 78.9 \rpm 1.2 & \textbf{82.0 \rpm 1.1}\\
C $\rightarrow$ R & 66.3 \rpm 1.5 & 63.6 \rpm 1.8 & 60.7 \rpm 1.1 & 65.8 \rpm 0.8 & \textbf{70.3 \rpm 0.8} & 73.0 \rpm 1.4 & 71.0 \rpm 3.6 & 67.6 \rpm 1.9 & 76.9 \rpm 3.0 & \textbf{80.3 \rpm 1.6}\\
P $\rightarrow$ A & 57.9 \rpm 0.5 & 51.2 \rpm 1.9 & 50.6 \rpm 1.4 & 55.9 \rpm 0.5 & \textbf{59.6 \rpm 1.0} & 71.5 \rpm 2.6 & 64.5 \rpm 4.0 & 65.6 \rpm 2.1 & 69.1 \rpm 0.5 & \textbf{74.3 \rpm 1.9}\\
P $\rightarrow$ C & 54.0 \rpm 2.1 & 52.2 \rpm 4.3 & 57.9 \rpm 0.8 & 56.1 \rpm 3.5 & \textbf{59.2 \rpm 0.4} & 59.1 \rpm 2.8 & 56.3 \rpm 4.4 & 63.1 \rpm 1.9 & 63.7 \rpm 0.9 & \textbf{69.3 \rpm 2.5}\\
P $\rightarrow$ R & 69.6 \rpm 1.3 & 65.9 \rpm 1.9 & 66.0 \rpm 0.1 & 68.5 \rpm 0.0 & \textbf{72.9 \rpm 0.3} & 81.2 \rpm 1.5 & 76.4 \rpm 0.7 & 75.8 \rpm 3.3 & 80.2 \rpm 2.6 & \textbf{84.3 \rpm 1.5}\\
R $\rightarrow$ A & 61.3 \rpm 0.9 & 56.7 \rpm 1.4 & 55.0 \rpm 1.2 & 57.6 \rpm 1.0 & \textbf{63.5 \rpm 2.2} & 73.0 \rpm 1.2 & 69.0 \rpm 4.3 & 68.8 \rpm 2.1 & 73.0 \rpm 2.9 & \textbf{75.5 \rpm 2.2}\\
R $\rightarrow$ C & 56.4 \rpm 0.9 & 57.0 \rpm 2.3 & \textbf{59.1 \rpm 2.7} & 57.5 \rpm 1.1 & 59.0 \rpm 2.4 & 62.9 \rpm 2.6 & 56.7 \rpm 6.3 & 61.4 \rpm 2.5 & 66.8 \rpm 3.3 & \textbf{71.5 \rpm 3.5}\\
R $\rightarrow$ P & 74.3 \rpm 1.4 & 70.0 \rpm 1.0 & 72.5 \rpm 1.4 & 73.9 \rpm 1.3 & \textbf{76.9 \rpm 0.4} & \textbf{84.0 \rpm 1.4} & 81.1 \rpm 0.7 & 79.5 \rpm 0.5 & 83.3 \rpm 0.8 & 83.3 \rpm 1.4\\ \midrule
Average & 63.2 & 60.3 & 60.6 & 63.1 & \textbf{66.9} & 72.9 & 68.7 & 69.7 & 74.4 & \textbf{77.4}\\
\midrule[0.3pt]\bottomrule[1pt]
\end{tabular}
\end{adjustbox}
\caption{Open-set and partial-set settings \label{tab: results_full_officehome_open_partial}}
\end{subtable}

\begin{subtable}{\textwidth}
\centering
\begin{adjustbox}{max width=0.55\columnwidth}
\begin{tabular}{c|*{5}{c}}
\toprule[1pt]\midrule[0.3pt]

\textbf{S $\rightarrow$ T}  & \multicolumn{5}{c}{\textbf{Open-partial}} \\ \cmidrule(lr){2-6}
& S + T & CDAC & PAC & AdaMatch & \makecell{Proposed}\\ \midrule
A $\rightarrow$ C & 54.6 \rpm 2.9 & 43.9 \rpm 9.8 & \textbf{58.5 \rpm 1.9} & 55.2 \rpm 1.0 & 55.5 \rpm 1.1\\
A $\rightarrow$ P & 70.3 \rpm 0.5 & 60.3 \rpm 8.8 & 68.3 \rpm 1.2 & 73.3 \rpm 0.9 & \textbf{75.9 \rpm 1.4}\\
A $\rightarrow$ R & 73.9 \rpm 1.5 & 62.2 \rpm 1.3 & 67.0 \rpm 2.3 & 71.4 \rpm 1.6 & \textbf{76.0 \rpm 0.9}\\
C $\rightarrow$ A & 58.7 \rpm 3.1 & 50.1 \rpm 2.4 & 51.5 \rpm 2.0 & 54.6 \rpm 3.0 & \textbf{61.6 \rpm 0.9}\\
C $\rightarrow$ P & 70.4 \rpm 1.0 & 66.4 \rpm 0.9 & 68.4 \rpm 2.7 & 72.2 \rpm 0.8 & \textbf{75.1 \rpm 2.2}\\
C $\rightarrow$ R & 69.1 \rpm 1.0 & 57.6 \rpm 1.4 & 62.0 \rpm 1.0 & 67.9 \rpm 1.3 & \textbf{73.7 \rpm 1.0}\\
P $\rightarrow$ A & 60.0 \rpm 2.5 & 53.7 \rpm 2.9 & 54.4 \rpm 1.6 & 58.8 \rpm 1.5 & \textbf{64.0 \rpm 1.2}\\
P $\rightarrow$ C & 52.1 \rpm 1.3 & 49.4 \rpm 1.9 & \textbf{55.7 \rpm 3.1} & 54.4 \rpm 1.0 & 53.6 \rpm 2.3\\
P $\rightarrow$ R & 70.4 \rpm 0.7 & 63.3 \rpm 4.2 & 65.8 \rpm 0.3 & 70.6 \rpm 1.3 & \textbf{74.8 \rpm 0.7}\\
R $\rightarrow$ A & 61.9 \rpm 3.7 & 52.6 \rpm 0.2 & 55.5 \rpm 1.8 & 59.4 \rpm 1.1 & \textbf{63.7 \rpm 2.6}\\
R $\rightarrow$ C & 54.3 \rpm 0.3 & 48.8 \rpm 7.0 & \textbf{58.6 \rpm 2.6} & 56.0 \rpm 2.3 & 56.6 \rpm 0.9\\
R $\rightarrow$ P & 74.3 \rpm 1.3 & 67.6 \rpm 2.2 & 70.1 \rpm 2.0 & 74.9 \rpm 0.7 & \textbf{78.0 \rpm 3.3}\\ \midrule
Average & 64.2 & 56.3 & 61.3 & 64.1 & \textbf{67.4}\\
\midrule[0.3pt]\bottomrule[1pt]
\end{tabular}
\end{adjustbox}
\caption{Open-partial settings \label{tab: results_full_officehome_openpartial}}
\end{subtable}

\caption{Office-Home: Target domain accuracy for each source (S) to target (T) pair, trained on ResNet-34 backbone. \label{tab: results_full_officehome}}

\end{table*}

\begin{table*}[tb]

\begin{subtable}{\textwidth}
\centering
\begin{adjustbox}{max width=\columnwidth}
\begin{tabular}{c|*{5}{c}|*{5}{c}}
\toprule[1pt]\midrule[0.3pt]

\textbf{S $\rightarrow$ T}  & \multicolumn{5}{c|}{\textbf{Closed-set w/o Label Distribution Shift}}           
                            & \multicolumn{5}{c}{\textbf{Closed-set w/ Label Distribution Shift}} \\ \cmidrule(lr){2-6} \cmidrule(lr){7-11}
& S + T & CDAC & PAC & AdaMatch & \makecell{Proposed} & S + T & CDAC & PAC & AdaMatch & \makecell{Proposed}\\ \midrule
C $\rightarrow$ P & 61.9 \rpm 1.0 & 66.9 \rpm 0.1 & 64.7 \rpm 0.4 & 62.9 \rpm 0.9 & \textbf{67.0 \rpm 1.4} & 55.3 \rpm 0.0 & 61.5 \rpm 0.2 & 60.3 \rpm 0.7 & 57.2 \rpm 1.4 & \textbf{63.7 \rpm 0.5}\\
C $\rightarrow$ R & 72.3 \rpm 0.5 & 75.5 \rpm 0.5 & 75.2 \rpm 1.1 & 75.3 \rpm 0.5 & \textbf{78.4 \rpm 0.6} & 69.0 \rpm 0.9 & 73.4 \rpm 1.3 & 71.5 \rpm 0.3 & 71.7 \rpm 1.1 & \textbf{75.4 \rpm 0.7}\\
C $\rightarrow$ S & 59.4 \rpm 0.3 & 65.8 \rpm 1.8 & 66.9 \rpm 0.7 & 63.4 \rpm 0.7 & \textbf{67.9 \rpm 1.2} & 52.2 \rpm 0.5 & 59.4 \rpm 1.1 & 58.1 \rpm 0.6 & 54.1 \rpm 0.9 & \textbf{61.6 \rpm 0.8}\\
P $\rightarrow$ C & 62.4 \rpm 0.6 & 70.7 \rpm 1.0 & 70.7 \rpm 0.3 & 65.0 \rpm 0.5 & \textbf{73.3 \rpm 0.2} & 54.9 \rpm 0.4 & 62.0 \rpm 1.8 & 65.5 \rpm 1.3 & 58.5 \rpm 1.1 & \textbf{66.6 \rpm 0.4}\\
P $\rightarrow$ R & 76.6 \rpm 0.4 & 77.8 \rpm 0.2 & 77.8 \rpm 0.5 & 76.7 \rpm 0.3 & \textbf{79.0 \rpm 0.2} & 73.0 \rpm 1.0 & 75.3 \rpm 0.6 & 73.3 \rpm 1.0 & 73.6 \rpm 0.6 & \textbf{76.1 \rpm 0.3}\\
P $\rightarrow$ S & 56.3 \rpm 1.3 & 64.9 \rpm 0.6 & 64.9 \rpm 1.2 & 60.7 \rpm 0.8 & \textbf{67.2 \rpm 1.5} & 49.5 \rpm 0.5 & 60.2 \rpm 1.0 & 59.3 \rpm 1.6 & 55.3 \rpm 0.3 & \textbf{61.5 \rpm 1.0}\\
R $\rightarrow$ C & 58.3 \rpm 0.2 & 67.7 \rpm 0.4 & 66.2 \rpm 2.9 & 63.6 \rpm 1.1 & \textbf{71.1 \rpm 0.4} & 54.6 \rpm 1.0 & 63.6 \rpm 0.9 & 64.5 \rpm 0.5 & 58.3 \rpm 1.3 & \textbf{66.7 \rpm 1.1}\\
R $\rightarrow$ P & 63.8 \rpm 0.4 & 66.9 \rpm 1.1 & 66.6 \rpm 0.1 & 64.3 \rpm 0.4 & \textbf{68.5 \rpm 0.7} & 60.8 \rpm 0.3 & 67.1 \rpm 0.5 & 65.4 \rpm 0.8 & 61.9 \rpm 1.6 & \textbf{67.6 \rpm 0.4}\\
R $\rightarrow$ S & 52.4 \rpm 1.6 & 62.5 \rpm 0.4 & 63.1 \rpm 0.8 & 58.5 \rpm 0.5 & \textbf{66.0 \rpm 0.3} & 47.9 \rpm 0.3 & 58.2 \rpm 0.5 & 58.2 \rpm 1.3 & 52.3 \rpm 1.7 & \textbf{60.1 \rpm 0.0}\\
S $\rightarrow$ C & 66.9 \rpm 0.7 & 72.7 \rpm 0.9 & 73.0 \rpm 0.2 & 69.0 \rpm 0.6 & \textbf{75.1 \rpm 1.5} & 61.3 \rpm 0.3 & 66.8 \rpm 0.5 & 67.4 \rpm 0.8 & 62.4 \rpm 1.1 & \textbf{69.5 \rpm 0.8}\\
S $\rightarrow$ P & 63.6 \rpm 0.4 & 67.6 \rpm 0.3 & 66.2 \rpm 0.8 & 65.3 \rpm 0.5 & \textbf{69.3 \rpm 0.2} & 59.4 \rpm 0.3 & 64.3 \rpm 0.7 & 62.6 \rpm 0.1 & 59.9 \rpm 0.9 & \textbf{64.6 \rpm 1.0}\\
S $\rightarrow$ R & 73.3 \rpm 0.6 & 77.0 \rpm 0.3 & 74.2 \rpm 1.0 & 75.7 \rpm 0.5 & \textbf{78.4 \rpm 0.5} & 68.1 \rpm 0.9 & 72.2 \rpm 0.5 & 68.6 \rpm 0.5 & 71.0 \rpm 0.8 & \textbf{74.0 \rpm 0.4}\\ \midrule
Average & 63.9 & 69.7 & 69.1 & 66.7 & \textbf{71.8} & 58.8 & 65.3 & 64.6 & 61.3 & \textbf{67.3}\\
\midrule[0.3pt]\bottomrule[1pt]
\end{tabular}
\end{adjustbox}
\caption{Closed set and class distribution shift settings \label{tab: results_full_domainnet_closed_cds}}
\end{subtable}

\begin{subtable}{\textwidth}
\centering
\begin{adjustbox}{max width=\columnwidth}
\begin{tabular}{c|*{5}{c}|*{5}{c}}
\toprule[1pt]\midrule[0.3pt]

\textbf{S $\rightarrow$ T}  & \multicolumn{5}{c|}{\textbf{Open-set}}           
                            & \multicolumn{5}{c}{\textbf{Partial-set}} \\ \cmidrule(lr){2-6} \cmidrule(lr){7-11}
& S + T & CDAC & PAC & AdaMatch & \makecell{Proposed} & S + T & CDAC & PAC & AdaMatch & \makecell{Proposed}\\ \midrule
C $\rightarrow$ P & 52.2 \rpm 0.4 & 50.0 \rpm 0.5 & 46.8 \rpm 1.6 & 50.2 \rpm 1.1 & \textbf{57.4 \rpm 0.7} & 72.6 \rpm 1.3 & 74.8 \rpm 0.8 & 75.2 \rpm 1.5 & 73.9 \rpm 0.7 & \textbf{78.5 \rpm 1.0}\\
C $\rightarrow$ R & 68.0 \rpm 1.1 & 65.7 \rpm 1.7 & 62.1 \rpm 1.1 & 65.4 \rpm 1.9 & \textbf{72.6 \rpm 0.9} & 81.8 \rpm 0.3 & 82.3 \rpm 1.2 & 83.7 \rpm 1.8 & 83.1 \rpm 1.3 & \textbf{85.9 \rpm 1.1}\\
C $\rightarrow$ S & 46.4 \rpm 1.6 & 44.8 \rpm 1.9 & 46.8 \rpm 0.4 & 47.4 \rpm 0.6 & \textbf{53.8 \rpm 0.4} & 68.5 \rpm 0.9 & 70.3 \rpm 2.0 & 73.7 \rpm 0.3 & 72.4 \rpm 0.4 & \textbf{76.0 \rpm 0.3}\\
P $\rightarrow$ C & 50.1 \rpm 1.5 & 49.0 \rpm 2.2 & 48.0 \rpm 1.5 & 50.5 \rpm 0.5 & \textbf{60.9 \rpm 0.1} & 68.7 \rpm 0.1 & 73.2 \rpm 2.8 & 78.6 \rpm 1.5 & 75.9 \rpm 1.1 & \textbf{81.6 \rpm 0.1}\\
P $\rightarrow$ R & 69.4 \rpm 1.2 & 67.2 \rpm 1.1 & 63.7 \rpm 0.7 & 64.9 \rpm 1.5 & \textbf{72.9 \rpm 1.5} & 83.6 \rpm 0.5 & 82.6 \rpm 1.1 & 84.6 \rpm 0.5 & 83.6 \rpm 2.0 & \textbf{86.1 \rpm 0.9}\\
P $\rightarrow$ S & 45.5 \rpm 0.5 & 43.7 \rpm 1.7 & 47.2 \rpm 0.8 & 46.8 \rpm 1.1 & \textbf{54.2 \rpm 0.8} & 64.0 \rpm 1.2 & 69.6 \rpm 0.1 & 72.9 \rpm 0.3 & 70.6 \rpm 0.6 & \textbf{74.7 \rpm 0.7}\\
R $\rightarrow$ C & 47.7 \rpm 1.4 & 47.3 \rpm 2.7 & 47.4 \rpm 0.7 & 48.8 \rpm 2.1 & \textbf{58.7 \rpm 1.0} & 64.9 \rpm 2.5 & 74.6 \rpm 0.7 & 78.0 \rpm 0.7 & 74.7 \rpm 1.7 & \textbf{79.9 \rpm 3.8}\\
R $\rightarrow$ P & 53.5 \rpm 0.9 & 48.6 \rpm 4.8 & 48.7 \rpm 0.5 & 49.6 \rpm 2.0 & \textbf{57.2 \rpm 0.9} & 75.2 \rpm 0.6 & 76.4 \rpm 0.4 & 76.9 \rpm 0.8 & 75.3 \rpm 1.0 & \textbf{79.4 \rpm 0.5}\\
R $\rightarrow$ S & 41.9 \rpm 0.5 & 41.7 \rpm 0.8 & 46.6 \rpm 0.9 & 45.7 \rpm 1.6 & \textbf{52.4 \rpm 1.0} & 60.3 \rpm 1.5 & 65.8 \rpm 1.5 & 71.9 \rpm 0.6 & 67.2 \rpm 1.0 & \textbf{72.4 \rpm 2.6}\\
S $\rightarrow$ C & 53.2 \rpm 0.8 & 50.0 \rpm 2.0 & 50.9 \rpm 1.1 & 51.3 \rpm 2.1 & \textbf{62.0 \rpm 1.1} & 73.9 \rpm 0.4 & 76.0 \rpm 0.7 & 80.4 \rpm 0.9 & 79.1 \rpm 0.8 & \textbf{83.8 \rpm 0.8}\\
S $\rightarrow$ P & 52.5 \rpm 0.4 & 49.5 \rpm 3.6 & 47.8 \rpm 1.1 & 51.0 \rpm 1.1 & \textbf{58.3 \rpm 0.6} & 75.3 \rpm 0.5 & 75.6 \rpm 0.9 & 76.6 \rpm 1.1 & 75.0 \rpm 0.7 & \textbf{78.5 \rpm 1.7}\\
S $\rightarrow$ R & 68.4 \rpm 1.1 & 67.7 \rpm 1.6 & 62.6 \rpm 0.7 & 65.5 \rpm 1.7 & \textbf{73.6 \rpm 0.9} & 82.2 \rpm 0.3 & 81.9 \rpm 1.8 & 82.3 \rpm 1.5 & 84.8 \rpm 1.5 & \textbf{87.2 \rpm 1.3}\\ \midrule
Average & 54.1 & 52.1 & 51.6 & 53.1 & \textbf{61.2} & 72.6 & 75.3 & 77.9 & 76.3 & \textbf{80.3}\\
\midrule[0.3pt]\bottomrule[1pt]
\end{tabular}
\end{adjustbox}
\caption{Open-set and partial-set settings \label{tab: results_full_domainnet_open_partial}}
\end{subtable}

\begin{subtable}{\textwidth}
\centering
\begin{adjustbox}{max width=0.55\columnwidth}
\begin{tabular}{c|*{5}{c}}
\toprule[1pt]\midrule[0.3pt]

\textbf{S $\rightarrow$ T}  & \multicolumn{5}{c}{\textbf{Open-partial}} \\ \cmidrule(lr){2-6}
& S + T & CDAC & PAC & AdaMatch & \makecell{Proposed}\\ \midrule
C $\rightarrow$ P & 52.4 \rpm 0.2 & 47.2 \rpm 4.3 & 46.9 \rpm 1.3 & 51.9 \rpm 0.9 & \textbf{60.8 \rpm 1.2}\\
C $\rightarrow$ R & 69.7 \rpm 0.5 & 53.4 \rpm 17.8 & 63.1 \rpm 3.2 & 67.7 \rpm 0.7 & \textbf{74.7 \rpm 0.3}\\
C $\rightarrow$ S & 44.2 \rpm 2.1 & 42.3 \rpm 1.9 & 47.0 \rpm 3.7 & 46.3 \rpm 2.4 & \textbf{53.3 \rpm 2.9}\\
P $\rightarrow$ C & 51.3 \rpm 1.6 & 47.4 \rpm 1.1 & 47.5 \rpm 2.1 & 50.0 \rpm 1.0 & \textbf{59.4 \rpm 2.9}\\
P $\rightarrow$ R & 70.7 \rpm 0.3 & 44.9 \rpm 19.2 & 65.0 \rpm 1.6 & 67.0 \rpm 1.2 & \textbf{76.4 \rpm 0.5}\\
P $\rightarrow$ S & 45.1 \rpm 2.3 & 42.1 \rpm 2.6 & 45.9 \rpm 1.7 & 46.6 \rpm 3.1 & \textbf{54.4 \rpm 2.0}\\
R $\rightarrow$ C & 50.7 \rpm 1.6 & 48.7 \rpm 1.4 & 47.6 \rpm 3.5 & 50.2 \rpm 1.8 & \textbf{59.9 \rpm 2.0}\\
R $\rightarrow$ P & 53.6 \rpm 0.4 & 26.0 \rpm 11.3 & 49.9 \rpm 1.0 & 51.2 \rpm 1.6 & \textbf{61.5 \rpm 0.2}\\
R $\rightarrow$ S & 42.4 \rpm 1.9 & 23.3 \rpm 9.7 & 45.4 \rpm 3.0 & 44.4 \rpm 3.0 & \textbf{52.2 \rpm 1.7}\\
S $\rightarrow$ C & 53.0 \rpm 1.9 & 49.9 \rpm 0.4 & 48.8 \rpm 1.2 & 51.1 \rpm 1.3 & \textbf{61.3 \rpm 1.4}\\
S $\rightarrow$ P & 51.2 \rpm 0.4 & 48.8 \rpm 1.8 & 46.8 \rpm 1.9 & 50.1 \rpm 0.9 & \textbf{60.8 \rpm 0.4}\\
S $\rightarrow$ R & 67.8 \rpm 0.5 & 49.7 \rpm 15.8 & 61.1 \rpm 3.3 & 67.4 \rpm 0.2 & \textbf{74.9 \rpm 0.6}\\ \midrule
Average & 54.4 & 43.7 & 51.2 & 53.7 & \textbf{62.5}\\
\midrule[0.3pt]\bottomrule[1pt]
\end{tabular}
\end{adjustbox}
\caption{Open-partial settings \label{tab: results_full_domainnet_openpartial}}
\end{subtable}

\caption{DomainNet-126:  Target domain accuracy for each source (S) to target (T) pair, trained on ResNet-34 backbone. \label{tab: results_full_domainnet}}

\end{table*}

\begin{table*}[tb]

\begin{subtable}{\textwidth}
\centering
\begin{adjustbox}{max width=0.8\columnwidth}
\begin{tabular}{c|*{4}{c}|*{4}{c}}
\toprule[1pt]\midrule[0.3pt]

\textbf{S $\rightarrow$ T}  & \multicolumn{4}{c|}{\textbf{Closed-set w/ Label Distribution Shift}}           
                            & \multicolumn{4}{c}{\textbf{Open-set}} \\ \cmidrule(lr){2-5} \cmidrule(lr){6-9}
& S + T & DANCE & UniOT & \makecell{Proposed} & S + T & DANCE & UniOT & \makecell{Proposed}\\ \midrule
C $\rightarrow$ P & 69.0 \rpm 0.3 & 68.6 \rpm 0.4 & 67.4 \rpm 0.2 & \textbf{69.7 \rpm 0.6} & 64.6 \rpm 0.3 & 63.3 \rpm 0.4 & 59.2 \rpm 0.2 & \textbf{65.8 \rpm 0.2}\\
C $\rightarrow$ R & 79.8 \rpm 0.4 & 79.6 \rpm 0.3 & 78.3 \rpm 0.1 & \textbf{80.3 \rpm 0.3} & 76.0 \rpm 0.5 & 74.9 \rpm 0.6 & 70.7 \rpm 0.6 & \textbf{78.5 \rpm 0.3}\\
C $\rightarrow$ S & 69.3 \rpm 0.1 & 69.6 \rpm 0.1 & 69.1 \rpm 0.1 & \textbf{70.0 \rpm 0.2} & \textbf{61.3 \rpm 0.2} & 60.1 \rpm 0.5 & 55.2 \rpm 0.4 & 61.2 \rpm 0.1\\
P $\rightarrow$ C & 75.4 \rpm 0.1 & 75.5 \rpm 0.2 & 73.9 \rpm 0.6 & \textbf{76.2 \rpm 0.3} & \textbf{71.0 \rpm 0.4} & 70.0 \rpm 0.4 & 65.5 \rpm 0.8 & 70.9 \rpm 0.5\\
P $\rightarrow$ R & 80.1 \rpm 0.1 & 79.7 \rpm 0.4 & 78.3 \rpm 0.2 & \textbf{80.5 \rpm 0.2} & 75.8 \rpm 0.8 & 75.0 \rpm 1.1 & 70.1 \rpm 0.9 & \textbf{78.2 \rpm 0.6}\\
P $\rightarrow$ S & 68.1 \rpm 0.3 & 68.7 \rpm 0.4 & 67.6 \rpm 0.1 & \textbf{69.3 \rpm 0.4} & \textbf{61.6 \rpm 0.3} & 60.3 \rpm 0.5 & 54.9 \rpm 0.3 & \textbf{61.6 \rpm 0.4}\\
R $\rightarrow$ C & 75.4 \rpm 0.2 & \textbf{76.2 \rpm 0.2} & 75.9 \rpm 0.2 & \textbf{76.2 \rpm 0.2} & \textbf{71.3 \rpm 0.3} & 70.1 \rpm 0.2 & 65.7 \rpm 0.7 & 71.1 \rpm 0.3\\
R $\rightarrow$ P & 70.3 \rpm 0.2 & 70.9 \rpm 0.1 & 70.1 \rpm 0.2 & \textbf{71.3 \rpm 0.2} & 64.6 \rpm 0.2 & 63.3 \rpm 0.2 & 57.8 \rpm 0.5 & \textbf{65.2 \rpm 0.3}\\
R $\rightarrow$ S & 67.8 \rpm 0.3 & 68.6 \rpm 0.5 & 68.1 \rpm 0.1 & \textbf{69.0 \rpm 0.3} & \textbf{61.3 \rpm 0.3} & 60.2 \rpm 0.9 & 55.2 \rpm 0.2 & 60.4 \rpm 0.3\\
S $\rightarrow$ C & 77.6 \rpm 0.1 & 77.9 \rpm 0.2 & 76.9 \rpm 0.5 & \textbf{78.0 \rpm 0.6} & 71.1 \rpm 0.5 & 69.9 \rpm 0.3 & 64.3 \rpm 0.3 & \textbf{71.7 \rpm 0.7}\\
S $\rightarrow$ P & 70.0 \rpm 0.2 & 70.2 \rpm 0.1 & 68.7 \rpm 0.5 & \textbf{71.1 \rpm 0.1} & 64.1 \rpm 0.3 & 63.2 \rpm 0.3 & 58.3 \rpm 0.4 & \textbf{65.1 \rpm 0.0}\\
S $\rightarrow$ R & 80.7 \rpm 0.2 & 80.0 \rpm 0.3 & 79.1 \rpm 0.2 & \textbf{81.0 \rpm 0.2} & 76.1 \rpm 0.7 & 75.1 \rpm 0.7 & 70.7 \rpm 0.7 & \textbf{78.3 \rpm 0.5}\\ \midrule
Average & 73.6 & 73.8 & 72.8 & \textbf{74.4} & 68.2 & 67.1 & 62.3 & \textbf{69.0}\\
\midrule[0.3pt]\bottomrule[1pt]
\end{tabular}
\end{adjustbox}
\caption{Label distribution shift and open-set settings}
\end{subtable}

\begin{subtable}{\textwidth}
\centering
\begin{adjustbox}{max width=0.8\columnwidth}
\begin{tabular}{c|*{4}{c}|*{4}{c}}
\toprule[1pt]\midrule[0.3pt]

\textbf{S $\rightarrow$ T}  & \multicolumn{4}{c|}{\textbf{Partial-set}}           
                            & \multicolumn{4}{c}{\textbf{Open-partial}} \\ \cmidrule(lr){2-5} \cmidrule(lr){6-9}
& S + T & DANCE & UniOT & \makecell{Proposed} & S + T & DANCE & UniOT & \makecell{Proposed}\\ \midrule
C $\rightarrow$ P & 75.8 \rpm 0.7 & 75.7 \rpm 0.7 & 70.8 \rpm 0.6 & \textbf{78.0 \rpm 0.6} & 68.3 \rpm 0.9 & 67.5 \rpm 0.9 & 64.2 \rpm 0.9 & \textbf{69.7 \rpm 0.8}\\
C $\rightarrow$ R & 84.3 \rpm 0.4 & 84.0 \rpm 0.1 & 79.8 \rpm 0.3 & \textbf{85.1 \rpm 0.2} & 78.2 \rpm 0.3 & 77.3 \rpm 0.3 & 73.4 \rpm 0.1 & \textbf{80.5 \rpm 0.4}\\
C $\rightarrow$ S & 78.3 \rpm 0.2 & 79.1 \rpm 0.4 & 74.0 \rpm 0.3 & \textbf{79.4 \rpm 0.2} & 61.0 \rpm 0.1 & 59.3 \rpm 0.5 & 54.9 \rpm 0.5 & \textbf{61.3 \rpm 0.1}\\
P $\rightarrow$ C & 84.1 \rpm 0.8 & 84.9 \rpm 1.0 & 79.7 \rpm 0.3 & \textbf{85.6 \rpm 0.7} & 73.0 \rpm 0.6 & 72.0 \rpm 0.6 & 66.8 \rpm 0.4 & \textbf{74.0 \rpm 0.4}\\
P $\rightarrow$ R & 84.1 \rpm 0.1 & 84.1 \rpm 0.2 & 79.0 \rpm 0.3 & \textbf{85.0 \rpm 0.2} & 78.1 \rpm 0.7 & 77.5 \rpm 0.4 & 71.8 \rpm 0.3 & \textbf{80.0 \rpm 0.5}\\
P $\rightarrow$ S & 77.8 \rpm 0.5 & 78.8 \rpm 0.5 & 72.8 \rpm 0.8 & \textbf{79.1 \rpm 0.4} & 61.1 \rpm 0.1 & 59.9 \rpm 0.4 & 54.2 \rpm 0.2 & \textbf{61.5 \rpm 0.6}\\
R $\rightarrow$ C & 85.1 \rpm 0.3 & 86.1 \rpm 0.5 & 81.3 \rpm 0.9 & \textbf{86.8 \rpm 0.1} & 73.3 \rpm 0.5 & 72.4 \rpm 0.7 & 66.7 \rpm 0.5 & \textbf{74.2 \rpm 0.6}\\
R $\rightarrow$ P & 76.8 \rpm 0.1 & 77.3 \rpm 0.4 & 71.8 \rpm 0.3 & \textbf{78.4 \rpm 0.3} & 69.1 \rpm 0.5 & 68.0 \rpm 0.7 & 64.0 \rpm 0.4 & \textbf{70.0 \rpm 0.7}\\
R $\rightarrow$ S & 77.8 \rpm 0.4 & 78.9 \rpm 0.5 & 75.2 \rpm 0.3 & \textbf{79.4 \rpm 0.3} & \textbf{61.0 \rpm 0.1} & 60.2 \rpm 0.2 & 54.7 \rpm 0.5 & 60.9 \rpm 0.8\\
S $\rightarrow$ C & 85.5 \rpm 0.1 & 86.5 \rpm 0.7 & 80.0 \rpm 0.7 & \textbf{86.9 \rpm 0.2} & 73.5 \rpm 0.6 & 72.4 \rpm 0.4 & 65.5 \rpm 0.9 & \textbf{75.1 \rpm 0.4}\\
S $\rightarrow$ P & 76.3 \rpm 0.7 & 77.1 \rpm 0.6 & 71.1 \rpm 0.7 & \textbf{78.6 \rpm 0.3} & 68.3 \rpm 0.7 & 67.3 \rpm 1.1 & 63.4 \rpm 0.4 & \textbf{69.8 \rpm 0.8}\\
S $\rightarrow$ R & 84.7 \rpm 0.2 & 84.2 \rpm 0.1 & 78.8 \rpm 0.2 & \textbf{85.4 \rpm 0.2} & 78.2 \rpm 0.5 & 77.5 \rpm 0.3 & 72.9 \rpm 0.1 & \textbf{80.6 \rpm 0.3}\\ \midrule
Average & 80.9 & 81.4 & 76.2 & \textbf{82.3} & 70.3 & 69.3 & 64.3 & \textbf{71.5}\\
\midrule[0.3pt]\bottomrule[1pt]
\end{tabular}
\end{adjustbox}
\caption{Partial-set and open-partial settings}
\end{subtable}

\caption{DomainNet-345:  Target domain accuracy for each source (S) to target (T) pair. Traing is performed with frozen DINOv2 encoder dino2\_vitl14 and learnable classifier. \label{tab: results_full_domainnet345_dino}}

\end{table*}

\begin{table*}[tb]

\begin{subtable}{\textwidth}
\centering
\begin{adjustbox}{max width=0.8\columnwidth}
\begin{tabular}{c|*{4}{c}|*{4}{c}}
\toprule[1pt]\midrule[0.3pt]

\textbf{S $\rightarrow$ T}  & \multicolumn{4}{c|}{\textbf{Closed-set w/ Label Distribution Shift}}           
                            & \multicolumn{4}{c}{\textbf{Open-set}} \\ \cmidrule(lr){2-5} \cmidrule(lr){6-9}
& S + T & DANCE & UniOT & \makecell{Proposed} & S + T & DANCE & UniOT & \makecell{Proposed}\\ \midrule
C $\rightarrow$ P & \textbf{73.1 \rpm 0.4} & 72.6 \rpm 0.5 & 72.7 \rpm 0.4 & 72.8 \rpm 0.6 & 64.0 \rpm 0.3 & 62.8 \rpm 0.7 & 57.5 \rpm 0.7 & \textbf{66.4 \rpm 0.4}\\
C $\rightarrow$ R & 84.9 \rpm 0.4 & 84.7 \rpm 0.3 & 84.0 \rpm 0.2 & \textbf{85.0 \rpm 0.4} & 78.1 \rpm 0.3 & 76.6 \rpm 0.4 & 70.8 \rpm 0.8 & \textbf{81.4 \rpm 0.6}\\
C $\rightarrow$ S & 72.7 \rpm 0.2 & 72.3 \rpm 0.7 & \textbf{73.3 \rpm 0.2} & 73.0 \rpm 0.4 & 61.2 \rpm 0.5 & 59.5 \rpm 0.5 & 53.9 \rpm 0.6 & \textbf{62.1 \rpm 0.7}\\
P $\rightarrow$ C & 78.3 \rpm 0.3 & 77.7 \rpm 0.3 & 77.8 \rpm 0.2 & \textbf{78.6 \rpm 0.5} & 72.1 \rpm 0.8 & 71.1 \rpm 0.6 & 66.3 \rpm 1.1 & \textbf{74.9 \rpm 0.8}\\
P $\rightarrow$ R & 83.8 \rpm 0.2 & 83.5 \rpm 0.3 & 82.8 \rpm 0.1 & \textbf{84.0 \rpm 0.2} & 77.6 \rpm 0.4 & 76.1 \rpm 0.6 & 69.6 \rpm 0.9 & \textbf{81.0 \rpm 0.2}\\
P $\rightarrow$ S & 70.9 \rpm 0.7 & 70.6 \rpm 0.4 & \textbf{71.3 \rpm 0.6} & 71.2 \rpm 0.4 & 60.9 \rpm 1.0 & 59.6 \rpm 1.2 & 53.3 \rpm 0.7 & \textbf{62.2 \rpm 0.8}\\
R $\rightarrow$ C & 79.6 \rpm 0.4 & 79.4 \rpm 0.2 & \textbf{80.7 \rpm 0.1} & 79.9 \rpm 0.2 & 72.9 \rpm 0.6 & 71.1 \rpm 0.8 & 66.4 \rpm 0.9 & \textbf{75.4 \rpm 0.8}\\
R $\rightarrow$ P & 73.9 \rpm 0.1 & 73.3 \rpm 0.2 & \textbf{74.7 \rpm 0.2} & 73.9 \rpm 0.2 & 63.7 \rpm 0.5 & 62.4 \rpm 0.3 & 56.1 \rpm 0.4 & \textbf{65.8 \rpm 0.4}\\
R $\rightarrow$ S & 71.6 \rpm 0.8 & 71.5 \rpm 0.2 & \textbf{73.0 \rpm 0.2} & 72.1 \rpm 0.4 & 61.5 \rpm 0.8 & 60.3 \rpm 1.1 & 54.1 \rpm 0.2 & \textbf{61.8 \rpm 1.1}\\
S $\rightarrow$ C & 80.2 \rpm 0.2 & 80.0 \rpm 0.5 & \textbf{80.9 \rpm 0.2} & 80.8 \rpm 0.2 & 73.0 \rpm 0.5 & 71.5 \rpm 0.7 & 65.8 \rpm 1.0 & \textbf{75.2 \rpm 0.6}\\
S $\rightarrow$ P & 73.9 \rpm 0.2 & 73.4 \rpm 0.3 & 73.6 \rpm 0.4 & \textbf{74.0 \rpm 0.3} & 63.8 \rpm 0.3 & 62.2 \rpm 0.4 & 57.0 \rpm 0.4 & \textbf{66.0 \rpm 0.5}\\
S $\rightarrow$ R & 85.1 \rpm 0.3 & 85.1 \rpm 0.2 & 84.0 \rpm 0.2 & \textbf{85.4 \rpm 0.2} & 77.9 \rpm 0.4 & 76.4 \rpm 0.5 & 70.7 \rpm 0.9 & \textbf{81.2 \rpm 0.7}\\ \midrule
Average & 77.3 & 77.0 & 77.4 & \textbf{77.5} & 68.9 & 67.5 & 61.8 & \textbf{71.1}\\
\midrule[0.3pt]\bottomrule[1pt]
\end{tabular}
\end{adjustbox}
\caption{Label distribution shift and open-set settings}
\end{subtable}

\begin{subtable}{\textwidth}
\centering
\begin{adjustbox}{max width=0.8\columnwidth}
\begin{tabular}{c|*{4}{c}|*{4}{c}}
\toprule[1pt]\midrule[0.3pt]

\textbf{S $\rightarrow$ T}  & \multicolumn{4}{c|}{\textbf{Partial-set}}           
                            & \multicolumn{4}{c}{\textbf{Open-partial}} \\ \cmidrule(lr){2-5} \cmidrule(lr){6-9}
& S + T & DANCE & UniOT & \makecell{Proposed} & S + T & DANCE & UniOT & \makecell{Proposed}\\ \midrule
C $\rightarrow$ P & 81.2 \rpm 0.3 & 80.1 \rpm 0.2 & 79.2 \rpm 0.6 & \textbf{81.5 \rpm 0.3} & 68.7 \rpm 0.7 & 66.6 \rpm 0.9 & 64.4 \rpm 0.4 & \textbf{71.2 \rpm 0.9}\\
C $\rightarrow$ R & 89.1 \rpm 0.2 & 89.1 \rpm 0.3 & 86.7 \rpm 0.1 & \textbf{89.2 \rpm 0.1} & 80.3 \rpm 0.4 & 78.4 \rpm 0.3 & 74.3 \rpm 0.5 & \textbf{83.3 \rpm 0.3}\\
C $\rightarrow$ S & 80.5 \rpm 0.5 & \textbf{80.7 \rpm 0.3} & 78.0 \rpm 0.5 & 80.3 \rpm 0.0 & 61.4 \rpm 0.4 & 60.0 \rpm 0.3 & 54.6 \rpm 0.2 & \textbf{63.3 \rpm 0.4}\\
P $\rightarrow$ C & 86.1 \rpm 0.2 & 86.0 \rpm 0.3 & 83.5 \rpm 0.7 & \textbf{86.2 \rpm 0.5} & 74.5 \rpm 0.8 & 72.8 \rpm 0.8 & 68.3 \rpm 0.3 & \textbf{76.9 \rpm 0.6}\\
P $\rightarrow$ R & 88.0 \rpm 0.4 & 88.2 \rpm 0.4 & 85.0 \rpm 0.4 & \textbf{88.5 \rpm 0.5} & 79.5 \rpm 0.3 & 77.9 \rpm 0.3 & 72.0 \rpm 0.6 & \textbf{82.7 \rpm 0.1}\\
P $\rightarrow$ S & 78.6 \rpm 1.1 & 79.4 \rpm 0.2 & 76.5 \rpm 0.3 & \textbf{79.9 \rpm 0.1} & 61.2 \rpm 0.5 & 59.9 \rpm 0.3 & 53.6 \rpm 0.8 & \textbf{63.2 \rpm 0.4}\\
R $\rightarrow$ C & 87.0 \rpm 0.2 & 87.3 \rpm 0.2 & 84.9 \rpm 0.3 & \textbf{87.6 \rpm 0.5} & 75.1 \rpm 0.7 & 73.5 \rpm 0.4 & 67.8 \rpm 0.7 & \textbf{77.0 \rpm 1.1}\\
R $\rightarrow$ P & \textbf{80.8 \rpm 0.3} & 80.1 \rpm 0.4 & 77.7 \rpm 0.7 & 80.5 \rpm 0.7 & 68.7 \rpm 0.5 & 67.3 \rpm 0.8 & 62.3 \rpm 0.4 & \textbf{71.1 \rpm 1.0}\\
R $\rightarrow$ S & \textbf{80.4 \rpm 0.5} & 80.2 \rpm 0.4 & 78.8 \rpm 0.3 & \textbf{80.4 \rpm 0.1} & 61.4 \rpm 0.8 & 60.1 \rpm 0.2 & 54.7 \rpm 0.6 & \textbf{63.2 \rpm 0.8}\\
S $\rightarrow$ C & 87.4 \rpm 0.2 & \textbf{87.6 \rpm 0.1} & 83.9 \rpm 0.5 & 87.5 \rpm 0.6 & 74.7 \rpm 1.2 & 73.4 \rpm 0.9 & 67.3 \rpm 0.4 & \textbf{76.9 \rpm 0.8}\\
S $\rightarrow$ P & 81.8 \rpm 0.4 & 81.5 \rpm 0.7 & 78.7 \rpm 0.4 & \textbf{82.3 \rpm 0.3} & 68.2 \rpm 0.8 & 66.4 \rpm 0.5 & 63.5 \rpm 0.5 & \textbf{71.3 \rpm 0.5}\\
S $\rightarrow$ R & 88.7 \rpm 0.1 & 89.0 \rpm 0.2 & 86.3 \rpm 0.1 & \textbf{89.3 \rpm 0.3} & 80.0 \rpm 0.4 & 78.6 \rpm 0.3 & 74.0 \rpm 0.6 & \textbf{83.5 \rpm 0.2}\\ \midrule
Average & 84.1 & 84.1 & 81.6 & \textbf{84.4} & 71.1 & 69.6 & 64.7 & \textbf{73.7}\\
\midrule[0.3pt]\bottomrule[1pt]
\end{tabular}
\end{adjustbox}
\caption{Partial-set and open-partial settings}
\end{subtable}

\caption{DomainNet-345:  Target domain accuracy for each source (S) to target (T) pair. Traing is performed with frozen CLIP encoder ViT-L/14@336px and learnable classifier. \label{tab: results_full_domainnet345_clip}}

\end{table*}

{

\end{document}